\documentclass[preprint,12pt]{elsarticle}

\usepackage{graphicx}
\usepackage{amsmath}
\usepackage{pgfplots}
\pgfplotsset{width=12cm,compat=1.9}
\usepackage{subfigure}

\journal{review}

\begin{document}

\begin{frontmatter}


\title{An Informal Introduction to Multiplet Neural Networks}



\author{Nathan E. Frick}

\address{Texas, United States}

\begin{abstract}
In the artificial neuron, I replace the dot product with the weighted Lehmer mean, which may emulate different cases of a generalized mean. The single neuron instance is replaced by a \textit{multiplet} of neurons which have the same averaging weights.  A group of outputs feed forward, in lieu of the single scalar. The generalization parameter is typically set to a different value for each neuron in the multiplet. 

I further extend the concept to a multiplet taken from the Gini mean.  Derivatives with respect to the weight parameters and with respect to the two generalization parameters are given.  

Some properties of the  network are investigated, showing the capacity to emulate the classical exclusive-or problem organically in two layers and perform some multiplication and division.  The multiplet network can instantiate truncated power series and variants, which can be used to approximate different functions, provided that parameters are constrained.

Moreover, a mean case slope score is derived that can facilitate a learning-rate novelty based on  homogeneity of the selected elements.  The multiplet neuron equation provides a way to segment regularization timeframes and approaches.

\end{abstract}


\end{frontmatter}


\section{Introduction}
\label{S:1}
The ubiquitous artificial neuron has been defined by the dot product of weights and input vector.  Alternative approaches have been introduced, such as the cosine distance\cite{DBLP:journals/corr/LuoZWY17}.   Others have shelved the dot product for geometric mean approaches\cite{Shiblee:2010:LGM:1841991.1842288}.  Generalized mean based neurons have been explored\cite{Yadav2006} with static generalization parameter.  Attempts to infuse logic into neural networks have been made\cite{DBLP:journals/corr/HuMLHX16}.  Methods for extraction of logical rules with the help of neural classifiers have been presented\cite{Duch1998}.  Weighted harmonic mean approaches have been introduced \cite{Zu2009} with triangular fuzzy variables.  Networks using parameterized ratios have been recently presented\cite{2020arXiv200506678Z}.  Here, I begin by introducing the use of the Lehmer mean \cite{GOULD1984611,havil_2003,ALZER1988c}, since it is differentiable, real monotonic, and amenable to algorithm optimization.

\subsection{The Weighted Lehmer Mean}

Considering for now input values that are positive, I assert that the weighted Lehmer mean\cite{bullen_2003}, with weight vector $\mathbf{w}$ (having elements $w_i$) and input vector $\mathbf{x}$ given by
\begin{equation}
\label{eq:wlm1}
\sum { w_i x_i^p } / \sum { w_i x_i^{p-1}}
\end{equation}
qualifies as an extension/generalization of the dot product, if we insist that we also denormalize by some gain $m$, as a type of reparameterization of the vector magnitude. See the literature for some similar reparameterization definitions.\cite{DBLP:journals/corr/SalimansK16}

When generalization parameter $p$ is varied, the Lehmer mean has cases where it acts as the maximum (when $p \rightarrow  \infty$), the standard mean, the geometric mean, the harmonic mean, or the minimum (when $p \rightarrow  -\infty$).  It also does not require any square root - or powers of $1/r$ - as generalized power means do.  We can investigate deprecating $\infty$ for a large enough magnitude number (e.g. $p=8$) for computational purposes.


\subsection{Definition and Derivative of the Lehmer Multiplet Neuron}

I define a \textit{multiplet} of neurons as a group of neurons in the same layer having the same input vector instance $\mathbf{x}$ and membership selection weights $\mathbf{w}$, but with different generalization parameter $p$.  Each neuron in the multiplet can instantiate a different Lehmer mean case. The Lehmer multiplet neuron $M$ has definition

\begin{equation}
\label{eq:lehmer_neuron}
  b + m {\sum { w_i x_i^p } / \sum { w_i x_i^{p-1}}}
\end{equation}
and the $w_i$ should have generally non-zero positive values.\footnote{Weight parameters are not all to be regarded as basis vector elements, in that the $w_i$ may be regarded as selectors.}  However, $m$ has no such requirement.  We can allow m and b within the multiplet, so that we can write
\begin{equation}
\label{eq:lehmer_neuron_m_by_j}
 M_j(\mathbf{x}) =  b_j + m_j {\sum_i { w_i x_i^{p_j} } / \sum_i { w_i x_i^{p_j-1}}} 
\end{equation}
where this is the $jth$ neuron in the multiplet having input vector elements $x_i$.  Note that $b_j$ is not relegated to remain a static layer offset, but may be function of a layer baseline interval. 

The total number of parameters $\phi_n$ in each Lehmer neuron multiplet is 
\begin{equation}
\label{eq:lehmer_neuron_num_params}
  \phi_n = n + 3 \psi
\end{equation}
where input element vector length is given by $n$, three is from the three other parameters $b_j,m_j,p_j$ in each neuron, and the number of neurons in the multiplet is given by $\psi$.

\subsubsection{Derivatives of interest for the Lehmer neuron}
The derivative with respect to the weight $w_k$ is
\begin{equation}
\label{eq:_neuron_derivative}
m \frac {x_k^p \sum { w_i x_i^{p-1} } - x_k^{p-1} \sum { w_i x_i^{p} }} {[{\sum { w_i x_i^{p-1}}]}^2}
\end{equation}
which can be rewritten for optimization in terms of the original numerator sum $N=\sum { w_i x_i^p}$ and denominator $D=\sum { w_i x_i^{p-1}}$ as
\begin{equation}
\label{eq:_neuron_der_nd}
\frac{\partial }{\partial w_k} ( b +  m { \frac{N}{D}}) = m \frac {D x_k^p - N x_k^{p-1}}  {D^2}
\end{equation}
which involves the input vector element corresponding to the weight.  For some powers $p$, this derivative can have a small value. The derivative with respect to $p$ - should it be needed - may be stated as
\begin{equation}
\label{eq:_neuron_w_r_to_p}
\frac{\partial }{\partial p} ( b +  m { \frac{N}{D}}) = m {\frac {D \sum {w_i x_i^p ln(x_i) } - N \sum { w_i x_i^{p-1} ln(x_i)} } {D^2}}
\end{equation}
which requires calculation of the natural logarithm for each element in the input vector $\mathbf{x}$. For powers $p > 7$ and about $p < -3$, this derivative (\ref{eq:_neuron_w_r_to_p}) is small.

\subsection{Lehmer Multiplet Configuration}
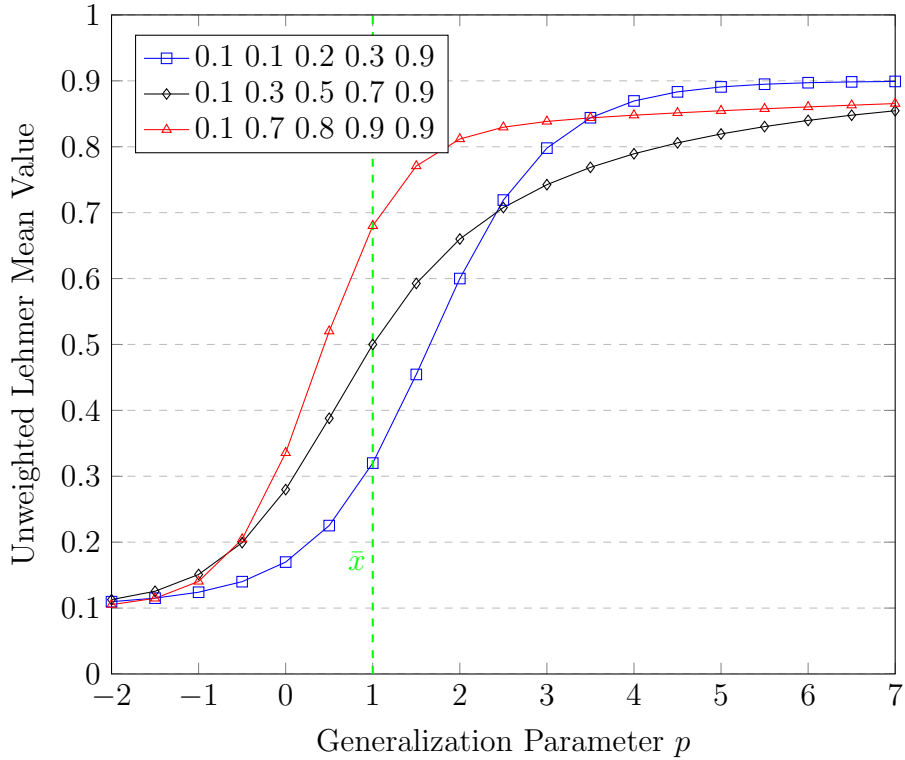
\begin{figure}[ht]

\begin{tikzpicture}
\label{tikzpicture:distcurves_pic1}
\begin{axis}[
    title={},
    xlabel={Generalization Parameter $p$ },
    ylabel={Unweighted Lehmer Mean Value},
    xmin=-2, xmax=7,
    ymin=0, ymax=1,
    legend pos=north west,
    ymajorgrids=true,
    grid style=dashed,
]
 
\addplot[
    color=blue,
    mark=square,
    ]
    coordinates {
    (-9.0,0.10005049666433538)(-8.5,0.1000719362545176)(-8.0,0.10010263497201674)(-7.5,0.10014670056172885)(-7.0,0.10021013693102941)(-6.5,0.1003017672875306)(-6.0,0.1004346361561056)(-5.5,0.10062815658375937)(-5.0,0.1009114303509944)(-4.5,0.10132844577513381)(-4.0,0.10194636339691726)(-3.5,0.10286911181019881)(-3.0,0.10426075978423045)(-2.5,0.1063885491651322)(-2.0,0.10970912261234843)(-1.5,0.11505533406997946)(-1.0,0.12405721716514956)(-0.5,0.14005937419080108)(0.0,0.169811320754717)(0.5,0.22517237058906625)(1.0,0.32)(1.5,0.4544976429204268)(2.0,0.6)(2.5,0.7191042133952582)(3.0,0.7979166666666667)(3.5,0.8439882255559265)(4.0,0.8694516971279372)(4.5,0.8832765925501687)(5.0,0.8907807807807809)(5.5,0.8948802270937568)(6.0,0.8971378484981288)(6.5,0.8983908388303881)(7.0,0.8990910047422572)(7.5,0.899484510220254)(8.0,0.8997067259929711)(8.5,0.8998327087708499)(9.0,0.8999043655875916)

    };
    
\addplot[
    color=black,
    mark=diamond,
    ]
    coordinates {
(-9.0,0.10000343027262303)(-8.5,0.10000596420757643)(-8.0,0.10001038225311132)(-7.5,0.10001810132464233)(-7.0,0.10003162491511226)(-6.5,0.10005540426236607)(-6.0,0.10009741936332284)(-5.5,0.10017213125959475)(-5.0,0.10030611100512286)(-4.5,0.10054903938337224)(-4.0,0.10099580003705934)(-3.5,0.10183211751383577)(-3.0,0.10343116240300569)(-2.5,0.10656072054476193)(-2.0,0.11282165400967334)(-1.5,0.1255110829572967)(-1.0,0.15097174573717323)(-0.5,0.19927251147718508)(0.0,0.2797513321492007)(0.5,0.3879534630173396)(1.0,0.5)(1.5,0.5925901760430594)(2.0,0.6599999999999999)(2.5,0.7077185393829573)(3.0,0.7424242424242425)(3.5,0.7687083364066902)(4.0,0.7893061224489796)(4.5,0.8058494624254111)(5.0,0.8193711862653842)(5.5,0.8305699397219092)(6.0,0.8399431997475545)(6.5,0.8478576051678157)(7.0,0.8545897857824464)(7.5,0.8603522440664422)(8.0,0.8653107665525139)(8.5,0.86959653777391)(9.0,0.873314731747992)

    };
    
\addplot[
    color=red,
    mark=triangle,
    ]
    coordinates {
(-9.0,0.1000000032348812)(-8.5,0.10000000884033824)(-8.0,0.10000002421384452)(-7.5,0.10000006647953955)(-7.0,0.10000018297203253)(-6.5,0.10000050488711208)(-6.0,0.10000139686272475)(-5.5,0.10000387521463538)(-5.0,0.10001078071288307)(-4.5,0.10003007653067861)(-4.0,0.10008414743138706)(-3.5,0.10023608037723532)(-3.0,0.10066401849182484)(-2.5,0.10187106699833373)(-2.0,0.10527112191481222)(-1.5,0.11476256186710272)(-1.0,0.1404775071351428)(-0.5,0.2045773155973031)(0.0,0.3355525965379494)(0.5,0.5201509007486885)(1.0,0.6799999999999999)(1.5,0.7707736158351047)(2.0,0.8117647058823529)(2.5,0.829620831158297)(3.0,0.8384057971014494)(3.5,0.8437978188681857)(4.0,0.8478824546240277)(4.5,0.8513993246280624)(5.0,0.8546075433231398)(5.5,0.8575983452230496)(6.0,0.8604053103045194)(6.5,0.8630427538116935)(7.0,0.8655189384172193)(7.5,0.867840521201906)(8.0,0.8700139685362765)(8.5,0.8720459338620107)(9.0,0.8739432907082089)

    };
    \legend{0.1 0.1 0.2 0.3 0.9, 0.1 0.3 0.5 0.7 0.9, 0.1 0.7 0.8 0.9 0.9}
    \addplot[
    color=green,thick,dashed,
    mark=none,
    ]
    coordinates { (1.0,0.0)(1.0,1.0) };
\end{axis}
    \node [right,color=green] at (3,1.5) {$\bar{x}$};

\end{tikzpicture}

\caption{Three Examples of 5 Elements Each, from Flat and Skewed Distributions}
\label{figure:distcurves}
\end{figure}

The elements of input vector $\mathbf{x}$ may have been generated from normal, skewed, or unusual distributions.  An examination of figure \ref{figure:distcurves} shows calculation of the Lehmer mean for three different groups of five values, from zero to one. 

\begin{table}

\centering
\begin{tabular}{l l l}
\hline
\textbf{Power $p$} & \textbf{Role} \\
\hline
$-3$ & Calculated Minimum \\
$-1$ & Post-Minimum \\
$0$  & Harmonic Mean  \\
$1$ & Arithmetic Mean  \\
$2$ & Contraharmonic Mean  \\
$3$ & Super-Contraharmonic Mean  \\
$5$ & Pre-Maximum  \\
$8$ & Calculated Maximum \\
\hline
\end{tabular}
\caption{Cases of the Lehmer Mean in an Eight Neuron Multiplet, with Integer Generalization Parameters}
\label{table:preferredcases8}
\end{table}

The graph also marks arithmetic mean. Table \ref{table:preferredcases8} shows a neuron octet at generalization parameters adjacent to the intersections shown, although this multiplet configuration is likely not optimal in practice, since it will have excessive co-dependence between outputs.

\section{The Perceptron Revisited}
\begin{figure}[ht]
\centering
\begin{subfigure}
  \centering
  \includegraphics[width=.4\linewidth]{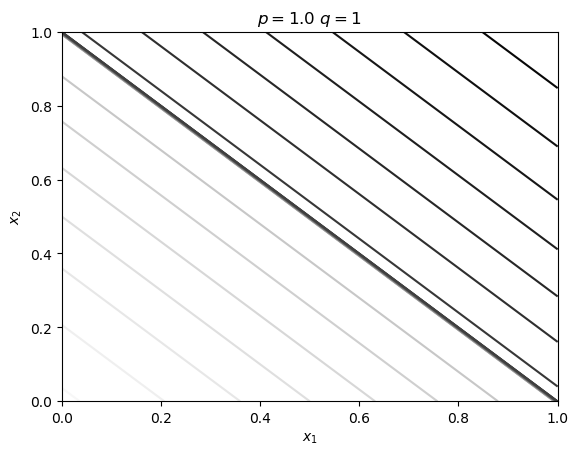}
\end{subfigure}
\begin{subfigure}
  \centering
  \includegraphics[width=.4\linewidth]{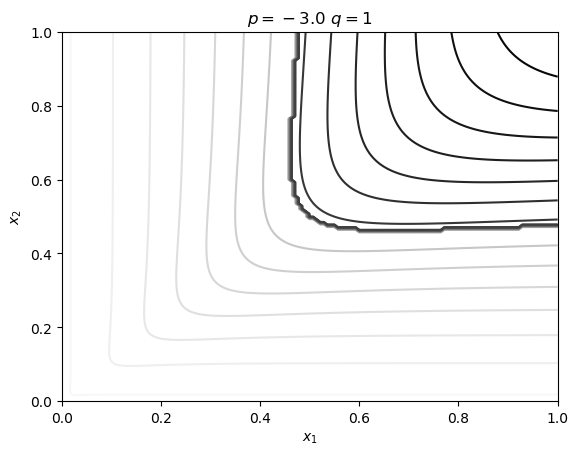}
\end{subfigure}
\centering
\begin{subfigure}
  \centering
  \includegraphics[width=.4\linewidth]{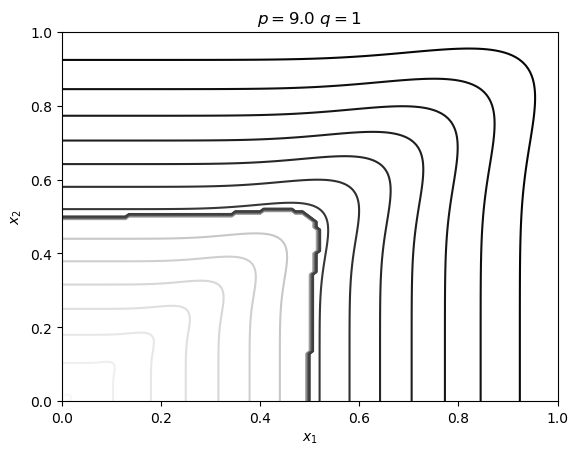}
\end{subfigure}
\begin{subfigure}
  \centering
  \includegraphics[width=.4\linewidth]{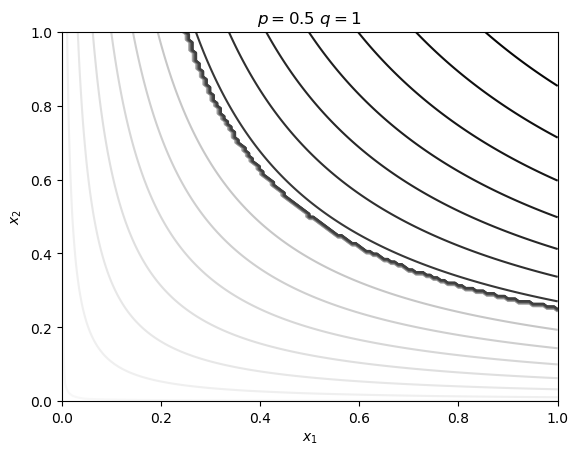}
\end{subfigure}
\caption{The Unweighted Multplet Perceptron, shown as Arithmetic Mean, Calculated Minimum, Calculated Maximum, and Geometric Mean}
\label{figure:perceptron_cases}
\end{figure}

The effect of the generalization parameter on the ubiquitous perceptron can be shown graphically.  Figure \ref{figure:perceptron_cases} illustrates how the linear classification line is modified nonlinearly for calculated maximum $p=9$ and calculated minimum $p=-3$ using a two element vector.  (I added a hyperbolic function to the surface to aid the illustration).   

\section{Properties}

I informally discuss some properties and capabilities of interest.  The universal function approximator argument may be found in the literature, classically\cite{ChernoffH47} and recently by Kidger and Lyons  \cite{KidgerLyons19} or by Molina, et al\cite{MolinaSK19}.

\subsection{Single Element Pass-through}
For any given multiplet, a single input vector element can pass through the layer, when all other $w_i$ are zero.  It can pass through unmodified (i.e. $m=1$, $b=0$), or it can be subjected to a linear transform by the values of $m$ and $b$.

\subsection{Affine Transformations and Reduction to the Dot Product}
As in classical networks, when generalization parameter $p=1$, affine transformations can occur.  This can be accomplished in one neuron.

In the dot product, when all coefficients of the first vector are positive, the multiplet neuron can be reduced to this dot product by simple scaling by $m$ of the normalized weights when $p=1$.  However, if we want to provide equivalence to the dot product with positive and negative coefficients, this must accomplished by varied values of $m_j$ and $w_i$ in more than one neuron multiplet and in two layers.  One multiplet must select elements (by using $w_i$) related to negative-valued $m$ and another multiplet must select the others.  Then, the final sum of the dot product terms must be accomplished by a neuron in the next layer.

\subsection{Measure of Independence of Neurons in a Multiplet}
\begin{figure}[ht]
\centering\includegraphics[width=0.7\linewidth]{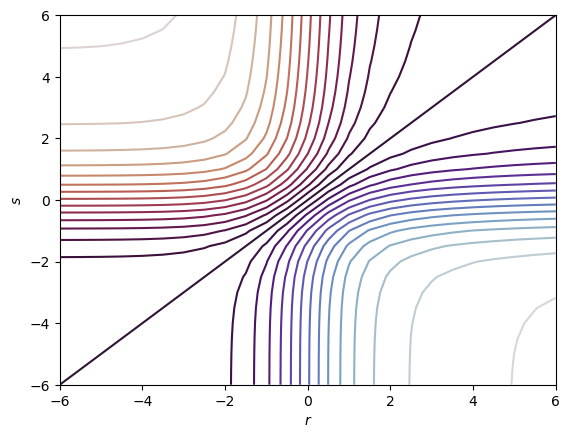}
\caption{The logarithm of co-dependence function $c(\mathbf{x})$, showing that generalization parameter $p$ cases of higher magnitude and opposite sign are most independent with same input $\mathbf{x}$ }
\label{figure:co-depend-2d}
\end{figure}

A lack of independence between neurons with different values of $p$ seems obvious, since it is a  function.  However, since linear independence is a topic of interest to the machine learning practitioner, it seems suitable to discuss.  A function $M_p$ can be said to be dependent in some way if
\begin{equation}
    \label{eq:lm_dependence_eqty}
    M_r - c(\mathbf{x}) M_s = 0
\end{equation}
where $r$ and $s$ represent non-identical values of $p$ and where $c(\mathbf{x})$ is some co-dependence factor.  Let us begin by ignoring $b$ and assuming $w_i$ are all identical such that
\begin{equation}
    \label{eq:lm_dependence_eqty2}
    c(\mathbf{x}) = \frac{\sum{x_i^r} \sum{x_i^{s-1}}} {\sum{x_i^s} \sum{x_i^{r-1}}}
\end{equation}
which will be exactly 1 when $r=s$.  As shown in figure \ref{figure:co-depend-2d}, the calculated maximum and minimum cases have the most independence from one another.

\subsection{Possible Numerical Precision Issues}

When a small number (e.g. $0.0001$) is squared, the numeric precision required - relative to a number such as $1.0$ - is not intractable with floating point representations.  The double precision IEEE 754 standard\cite{IEEE:754-2019} specifies 15 or 16 significant decimal digits.   So, adding $10^{-8}$ to $1.0$ is generally not a problem.

However, take $10^{-4}$ to the power 6, and it becomes an issue to keep enough significant digits.  Adding $10^{-24}$ to $1.0$ requires more precision than most systems typically use.  Alternatives are to use libraries, higher precision processors, or special techniques.

For complex numbers, a suitable example\cite{Abramowitz} is
\begin{equation}
    \label{eq:abram_compl_eg3_7_21}
     z^5 = x^5 - 10 x^3 y^2 + 5 x y^4 + i(5 x^4 y - 10 x^2 y^3 + y^5)
\end{equation}
where the real and imaginary parts (i.e. $x$ and $y$) are raised to powers that could potentially wreak havoc with floating point limitations.

\subsection{Spectral Noise in the Input Vector}

As an examination of how a noisy signal propagates in the multiplet network, we can assume an input that has a small, identical additive noise $\eta$ at each element of the input vector.  Each $x_i$ is part signal $u_i$ and part noise $\eta_i$, where $\eta_i$ is a  constant, with alternating sign, so that when $i=1$, $\eta_i$ is positive, and when $i=2$, $\eta_i$ is negative, etc.
\begin{equation}
    \label{eq:noise_identical_add}
     x_i = u_i + \eta_i 
\end{equation}
Then, substituting
\begin{equation}
    \label{eq:noise_identical_sum1}
     \sum{\frac{x_i^p}{x_i^{p-1}}} = \sum{\frac{(u_i + \eta_i)^p}{(u_i + \eta_i)^{p-1}}}
\end{equation}
For large signal relative to the error, the Laurent series expansion of this equation about $u_i = \infty$ has the form
\begin{equation}
    \label{eq:noise_identical_Laurent}
    O(u) + O(\eta_i) + \sum_{j=1}^{\infty} O(\eta_i^{j+1}/u^j)
\end{equation}
in which the terms in the sum tend to approach zero.  Thus, small alternating noise only affects the result on the order of the magnitude of the noise itself.  Analysis of other noise configurations or sources is left to the reader.

\subsection{Construction of Logical Connectives}

Using logical reference, we can investigate some basic properties of the multiplet neuron for positive input values.  First, if we introduce constant $T$ (e.g. 1.0) where we let logical complement transform $\neg$ be
\begin{equation}
    \label{eq:involutive_neg}
    \neg x_1 = T - x_1
\end{equation}
and let a soft conjunction $\land$ be
\begin{equation}
    \label{eq:def_conjunction_I}
   x_1 \land x_2 = \frac{x_1^{-3} + x_2^{-3}}{x_1^{-4} + x_2^{-4}}    
\end{equation}
and let a soft disjunction $\lor$ be
\begin{equation}
    \label{eq:def_disjunction_I}
  x_1 \lor x_2 = \frac{x_1^7 + x_2^7}{x_1^6 + x_2^6}    
\end{equation}
we can discuss some basic qualities for input values bounded by zero and $T$.
 
\subsubsection{Soft XOR Duet-Singlet}
From a simple, two element input vector $\mathbf{x}$, one composition of the continuous exclusive-or
\begin{equation}
\label{eq:xor_identity_logicsym}
 \chi = (x_1 \lor x_2) \land \neg (x_1 \land x_2)
\end{equation}
can be modeled using neurons in two layers. Here I define a "duet" is a multiplet of two neurons, having different $p$.  A "singlet" is defined as a multiplet of one neuron, typically with $p$ negative.

The duet is in the first layer, with the singlet in the second layer.  The first part $\sigma_1$ of the duet is (from $\ref{eq:def_disjunction_I}$ above)
\begin{equation}
    \label{eq:def__pxor_eg1a}
  \sigma_1 = \frac{x_1^7 + x_2^7}{x_1^6 + x_2^6}    
\end{equation}
and for the second part, let $b = T$ and $m = -1$
\begin{equation}
    \label{eq:def__pxor_eg1b}
  \sigma_2 = T - \frac{x_1^{-3} + x_2^{-3}}{x_1^{-4} + x_2^{-4}}   
\end{equation}
and the second layer singlet output $\chi$ is
\begin{equation}
    \label{eq:def__pxor_eg_layer2}
  \chi = \frac{\sigma_1^{-3} + \sigma_2^{-3}}{\sigma_1^{-4} + \sigma_2^{-4}}    
\end{equation}
which is the implementation of Equation \ref{eq:xor_identity_logicsym} and is a continuous soft-logic XOR accomplished in two layers without any activation function.

\begin{figure}[ht]

\begin{center}
\begin{tikzpicture}[x=0.75pt,y=0.75pt,yscale=-1,xscale=1]

\draw   (271.65,99.74) .. controls (271.65,83.43) and (291.43,70.21) .. (315.83,70.21) .. controls (340.22,70.21) and (360,83.43) .. (360,99.74) .. controls (360,116.04) and (340.22,129.26) .. (315.83,129.26) .. controls (291.43,129.26) and (271.65,116.04) .. (271.65,99.74) -- cycle ;
\draw    (230,130) -- (279.22,116.71) ;
\draw    (232.52,68) -- (275.44,85.71) ;

\draw   (271.65,189.74) .. controls (271.65,173.43) and (291.43,160.21) .. (315.83,160.21) .. controls (340.22,160.21) and (360,173.43) .. (360,189.74) .. controls (360,206.04) and (340.22,219.26) .. (315.83,219.26) .. controls (291.43,219.26) and (271.65,206.04) .. (271.65,189.74) -- cycle ;
\draw    (230,220) -- (279.22,206.71) ;
\draw    (232.52,158) -- (275.44,175.71) ;

\draw   (401.65,139.74) .. controls (401.65,123.43) and (421.43,110.21) .. (445.83,110.21) .. controls (470.22,110.21) and (490,123.43) .. (490,139.74) .. controls (490,156.04) and (470.22,169.26) .. (445.83,169.26) .. controls (421.43,169.26) and (401.65,156.04) .. (401.65,139.74) -- cycle ;
\draw    (360,189.74) -- (409.22,156.71) ;
\draw    (360,99.74) -- (405.44,125.71) ;

\draw (282,82) node [anchor=north west][inner sep=0.75pt]   [align=left] {{\small m=1,b=0}};
\draw (281,172) node [anchor=north west][inner sep=0.75pt]   [align=left] {{\small m=-1,b=1}};
\draw (294,102) node [anchor=north west][inner sep=0.75pt]   [align=left] {p = 7};
\draw (291,191) node [anchor=north west][inner sep=0.75pt]   [align=left] {p = -3};
\draw (412,122) node [anchor=north west][inner sep=0.75pt]   [align=left] {{\small m=1,b=0}};
\draw (421,141) node [anchor=north west][inner sep=0.75pt]   [align=left] {p = -3};
\draw (211,52.4) node [anchor=north west][inner sep=0.75pt]    {$x_{1}$};
\draw (211,142.4) node [anchor=north west][inner sep=0.75pt]    {$x_{1}$};
\draw (211,120.4) node [anchor=north west][inner sep=0.75pt]    {$x_{2}$};
\draw (211,210.4) node [anchor=north west][inner sep=0.75pt]    {$x_{2}$};

\end{tikzpicture}

\end{center}
\label{figure:diagram_duetsinglet_xor}
\caption{Diagram of the Unweighted XOR Duet-Singlet}
\end{figure}
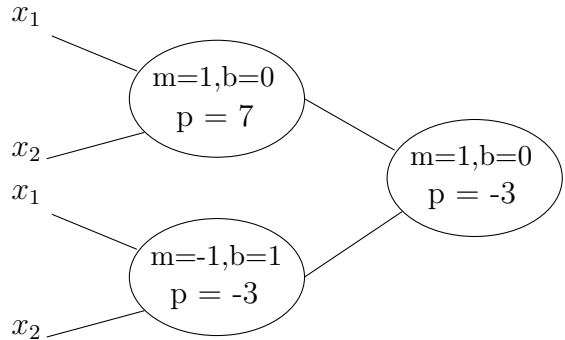

Table \ref{table:exxor} shows this calculation for non-zero values of real $x_1$ and $x_2$. With appropriate T value, the exclusive-or also works for values in an interval on the real axis, such as [1,2]. See Table \ref{table:exxor1to2}. However, this does not work in intervals that span zero (e.g. [-1,1]) since valid output is always near zero.  More accurate values are obtained when calculated minimum parameter $p$ is lower in $\sigma_2$.

\begin{table}
\centering
\begin{tabular}{l l l l l l}
\hline
$x_1$ & $x_2$ & \textbf{$\sigma_1$} & \textbf{$\sigma_2$} & \textbf{$\chi$} \\
\hline
0.01 & 0.01 & 0.01 & 0.99 & 0.01 \\
0.01 & 0.99 & 0.99 & 0.99 & 0.99 \\
0.99 & 0.01 & 0.99 & 0.99 & 0.99 \\
0.99 & 0.99 & 0.99 & 0.01 & 0.01 \\
\hline
\end{tabular}
\caption{A Calculation of the real XOR Duet-Singlet, with $b=T=1.0$}
\label{table:exxor}
\end{table}

\begin{table}
\centering
\begin{tabular}{l l l l l l}
\hline
$x_1$ & $x_2$ & \textbf{$\sigma_1$} & \textbf{$\sigma_2$} & \textbf{$\chi$} \\
\hline
1.0 & 1.0 & 1.0 & 2.0 & 1.05 \\
1.0 & 2.0 & 1.98 & 1.94 & 1.96 \\
2.0 & 1.0 & 1.98 & 1.94 & 1.96 \\
2.0 & 2.0 & 2.0 & 1.0 & 1.05 \\
\hline
\end{tabular}
\caption{A Calculation of the [1,2] XOR, with $b=T=3.0$ }
\label{table:exxor1to2}
\end{table}

\subsubsection{Complex Input and the Soft Exclusive-Or}
\begin{figure}[ht]
\centering\includegraphics[width=0.7\linewidth]{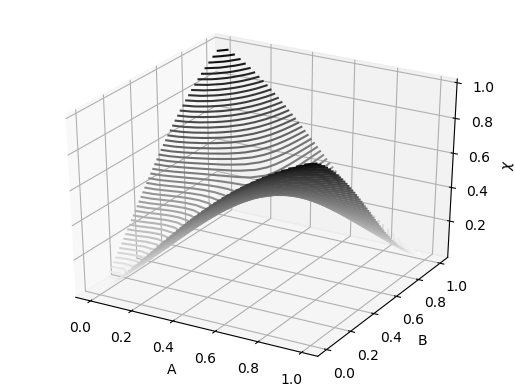}
\caption{Contiguous Real Surface of the Complex Soft XOR $\chi$, as in Table \ref{table:exxor_cmplx} }
\label{figure:xorsurf1_73}
\end{figure}

If the input values are allowed to be complex, a very small value $\varepsilon$ (e.g. 0.00001) may be assigned to the imaginary component.  We can recalculate the scenario given previously. 
\begin{table}

\centering
\begin{tabular}{l l l l l l}
\hline
$x_1$ & $x_2$ &  \textbf{$\chi$} \\
\hline
0.00 & 0.00 & 0.00\\
0.00 & 1.00 & 1.00 \\
1.00 & 0.00 & 1.00 \\
1.00 & 1.00 & -0.00 \\
\hline
\end{tabular}
\caption{Real Inputs and Outputs of the Complex XOR Duet-Singlet, with an initialization of $\varepsilon=0.000001$ for the imaginary component, showing equivalence with the classical binary XOR}
\label{table:exxor_cmplx}
\end{table}
With this initialization, we do not incur a \textit{divide by 0} exception, and we can use 0.0 and 1.0 exactly in the real component of the complex number and obtain (equivalently) the result of the classical XOR problem presentation!  
For a range of values in [0,1], a surface can be plotted, as shown\cite{Hunter:2007} in figure \ref{figure:xorsurf1_73}.
Note that for higher dimensions, compositions I and II will not really be equivalent to the formal XOR set definition.\footnote{Some compositions with more elements perhaps cannot be clearly defined.}  When a third input element $x_3$ is added, the $\chi$ surface "unwraps" and begins to tilt toward (or away) from the origin.  
\begin{figure}[ht]
\centering\includegraphics[width=0.7\linewidth]{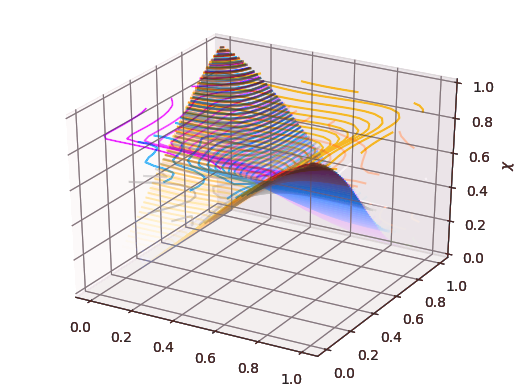}
\caption{The Unwrapping of $\chi$ when adding a third input element, where the orange surface is $x_3=0.17$ and magenta is $x_3=0.83$ }
\label{figure:xorsurf_tilting_color}
\end{figure}
The orange curves in the figure \ref{figure:xorsurf_tilting_color} show the surface from a smaller $x_3$ value of $0.17$, and the magenta curves show the surface from a larger $x_3$ value of $0.83$.

\subsubsection {Endpoint homogeneity}
\begin{figure}[ht]
\begin{tikzpicture}
\begin{axis}
[ 
enlargelimits=0.15,
legend style={at={(0.5,-0.1),col sep=2.5pt},
anchor=north,
legend columns=-1},
ybar,
xticklabels={,,},
xmajorticks=false,
bar width=7pt,]
\addplot coordinates{(1,0.85) (2,0.01) (3,0.1) (4,0.1) (5,0.4) };
\addplot coordinates{(1,0.9) (2,0.1) (3,0.85) (4,0.3) (5,0.5) };
\addplot coordinates{(1,0.94) (2,0.12) (3,0.9) (4,0.7) (5,0.6) };
\addplot coordinates{(1,0.99) (2,0.2) (3,0.94) (4,0.9) (5,0.7) };
\addplot coordinates{(1,0.91) (2,0.87) (3,0.09) (4,0.10) (5,0.44) };

\legend{$x_1$,$x_2$,$x_3$,$x_4$,composition II}
\end{axis}
\end{tikzpicture}
\caption{Data taken from XNOR Table \ref{table:calcs_nowt_exxnor}, illustrating high output (rightmost bar) for semi-homogeneous values near the interval ends  }
\label{figure:barcht_calcs_nowt_exxnor}
\end{figure}
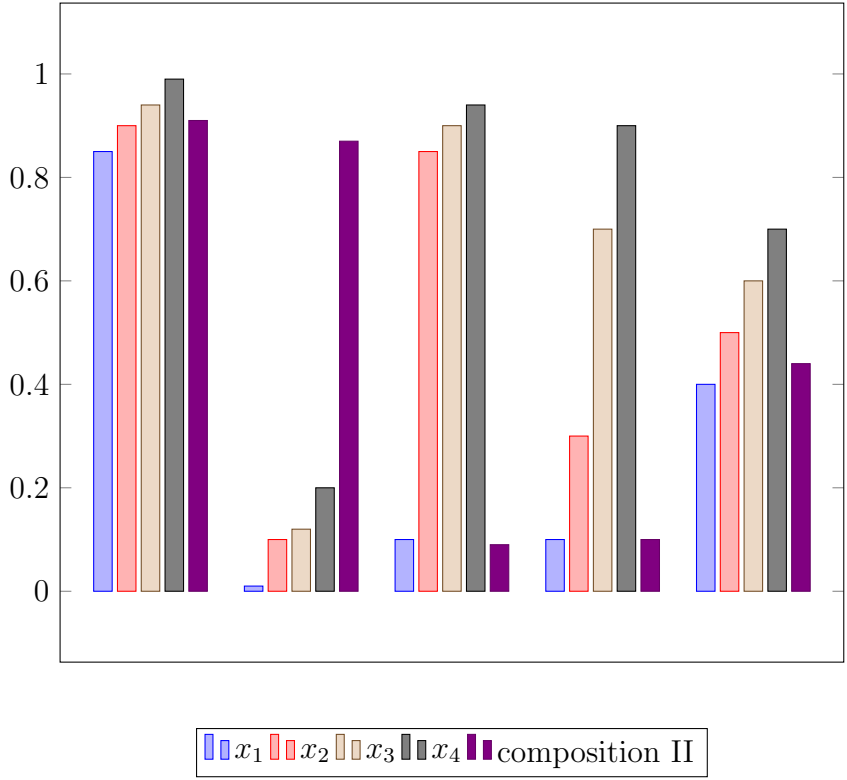

The XNOR is the logical complement of the XOR and can provide some measure of the homogeneity for values near zero and for values near $T$. See the rightmost bar in chart figure \ref{figure:barcht_calcs_nowt_exxnor}. Let $A$ be a subset containing $x_1$ and $x_3$ and let $B$ be a subset containing $x_2$ and $x_4$. We can write XNOR composition I as
\begin{equation}
\label{eq:me_logic_type_i}
 \neg((A \lor B) \land \neg (A \land B))
\end{equation}
and XNOR composition II as
\begin{equation}
\label{eq:me_logic_type_ii}
((A \land B) \lor (\neg A \land \neg B))
\end{equation}
which involves the preprocessing of every element in $A$ and $B$.  
\begin{table}

\centering
\begin{tabular}{l l l l l l}
\hline
\textbf{Input x} &  \textbf{I} &  \textbf{II}  \\
\hline
$0.85,0.9,0.94,0.99$ & $0.91$ & $0.91$  \\
$0.01,0.1,0.12,0.2$ & $0.81$ & $0.87$  \\
$0.1,0.85,0.9,0.94$ & $0.10$ & $0.09$  \\
$0.1,0.3,0.7,0.9$ & $0.15$ & $0.10$  \\
$0.4,0.5,0.6,0.7$ & $0.43$ & $0.44$  \\
\hline
\end{tabular}
\caption{ Output of the XNOR compositions I and II, Showing a Meta-Measure of Range-End Homogeneity}
\label{table:calcs_nowt_exxnor}
\end{table}

Table \ref{table:calcs_nowt_exxnor} shows the output for some $\mathbf{x}$. See also figure \ref{figure:barcht_calcs_nowt_exxnor}.  Note the output for the relatively homogeneous cluster near $T=1.0$ gives a value of $0.91$.  The output for a homogeneous cluster near zero also gives a high value of $0.87$.  For widely spaced values in the range, the  duet-singlet gives a lower value, such as $0.09$.  For values clustered in the middle (i.e. last row in the table), the output is $\approx{0.44}$, which is not descriptive in a range-end (e.g. one-hot value) interpretation of homogeneity.

\subsubsection {Input Interval Estimation}
A configuration exists whereby the interval estimate of an input vector can be output. Using small real constant $\epsilon$, a soft measure of the range of the input elements can be accomplished by 
\begin{equation}
\label{eq:composition_identity_range}
(\epsilon \lor X) \land (\epsilon \lor \neg X)
\end{equation}
The complemented elements of the input vector $\neg \mathbf{x}$ are used. See Table \ref{table:calcs_impl_duetsinglet}.  As with the XNOR Duet-Singlet, the output is not as descriptive when all the input values are near the midpoint value (i.e. row six in table).
\begin{table}

\centering
\begin{tabular}{l l l l l l}
\hline
\textbf{Input x} &  $(\epsilon \lor X)$ & $\mathbf{\neg x}$ & $(\epsilon \lor \neg X)$ & \textbf{Out}\\
\hline
$\epsilon,0.01,0.1,0.12,0.2$ & $0.20$ & $\epsilon,0.99,0.9,0.88,0.8$ & $0.93$ & $0.20$\\
$\epsilon,0.8,0.85,0.9,0.95$ & $0.90$ & $\epsilon,0.2,0.15,0.1,0.05$ & $0.20$ & $0.20$ \\
$\epsilon,0.05,0.75,0.9,0.95$ & $0.92$ & $\epsilon,0.95,0.25,0.1,0.05$ & $0.95$ & $0.93$ \\
$\epsilon,0.5,0.8,0.9,0.99$ & $0.94$ & $\epsilon,0.5,0.2,0.1,0.01$ & $0.50$ & $0.53$ \\
$\epsilon,0.1,0.2,0.3,0.4$ & $0.39$ & $\epsilon,0.9,0.8,0.7,0.6$ & $0.85$ & $0.41$ \\
$\epsilon,0.4,0.5,0.55,0.6$ & $0.57$ & $\epsilon,0.6,0.5,0.45,0.4$ & $0.57$ & $\mathbf{0.57}$ \\
\hline
\end{tabular}
\caption{ Output of Equation \ref{eq:composition_identity_range} with Real Constant $\epsilon = 0.0001$ and $p=9,9,-3$, Showing a Soft Measure of the Input Interval }
\label{table:calcs_impl_duetsinglet}
\end{table}

\section{Small Weights and the Disqualification of Input Vector Elements}

\begin{figure}[ht]
\centering\includegraphics[height=0.7\linewidth]{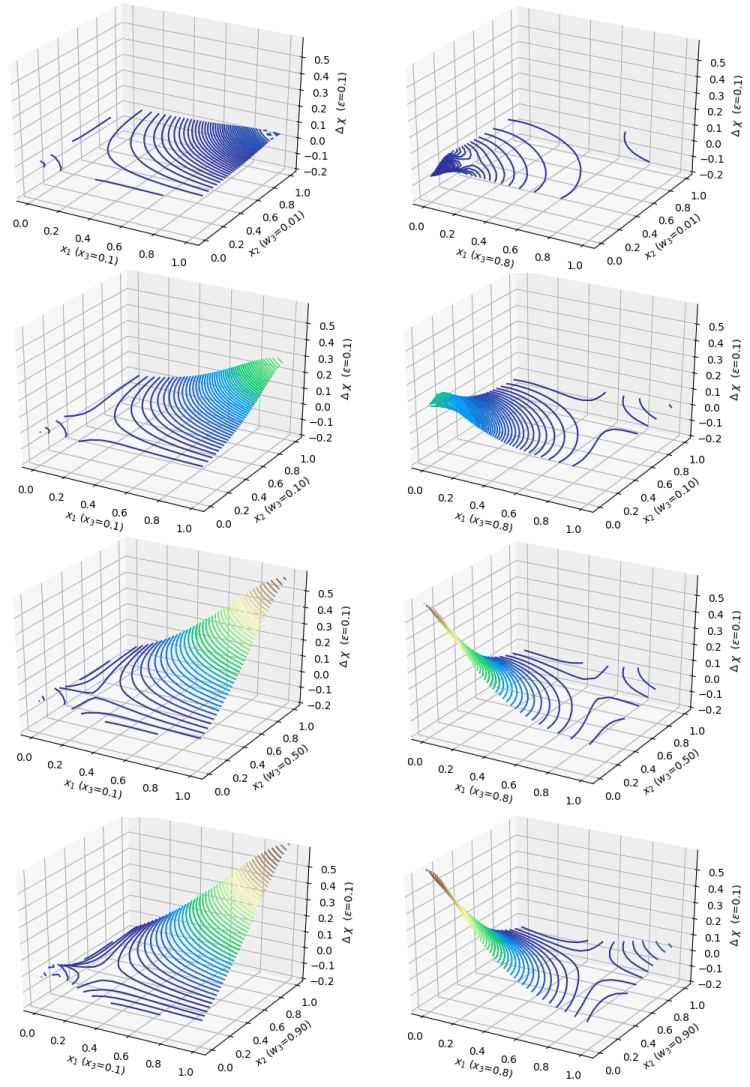}
\caption{Nonlinear sequence of weight $w_3$ values $0.01,0.1,0.5,0.9$ (top to bottom) used 
in calculating the $\chi$ surface change $\Delta$ when adding a constant third element $x_3$ of $0.1$ (on left) and $0.8$ (on right), showing a progression of deformation}
\label{figure:d20_xor_del8}
\end{figure}

What weight values $w_i$ will it take to essentially remove an input element $x_i$ from the Lehmer mean?  In the classical dot product, it was straightforward to dis-accentuate or disqualify a vector element with a small weight value (i.e. 0.1) relative to the others.  Here, figure \ref{figure:d20_xor_del8}  ($\varepsilon=0.1$) shows that the disqualification of a third element from soft XOR $\chi$ is certainly not linear.

A weight value of much less than the imaginary component constant $\varepsilon$ will disqualify the input element, as is desired.  A weight $w_3$ of $0.5$ and a weight $w_3$ of $0.9$ show a very similar $\Delta \chi$ surface.
It is possible to replace the weight terms by some function of the weights
\begin{equation}
\label{eq:lehmer_neuron_w_fx}
 b + m {\sum { f(w_i) x_i^p } / \sum { f(w_i) x_i^{p-1}}}
\end{equation}
If that function is to raise the weights to a power L, such that $M_p$ is now
\begin{equation}
\label{eq:lehmer_neuron_w_L}
 b + m {\sum { w_i^L x_i^p } / \sum { w_i^L x_i^{p-1}}}
\end{equation}
where $L$ would be a hyperparameter, low weight values would be made very small. ($L$ would be set to one of the higher values of $p$ in the multiplet.  For example, if the multiplets are defined from $p=-2$ to $p=3$, let $L=3$.)  The derivative with respect to $w_k$ in terms of numerator sum $N=\sum { w_i^L x_i^p}$ and denominator $D=\sum { w_i^L x_i^{p-1}}$  is
\begin{equation}
\label{eq:_neuron_der_nd_w}
\frac{\partial }{\partial w_k} ( b +  m { \frac{N}{D}}) = m L w_k^{L-1} \frac { D x_k^p - N x_k^{p-1}}  {D^2}
\end{equation}
which would supersede Equation \ref{eq:_neuron_der_nd}.  Other weight constraints are discussed later.

\section {Preliminary Engineered Tests}

\subsection {A Nearest Neighbor Search Test Using a Single Layer}

Using real input values, I preprocess the MNIST LeCun dataset\cite{LeCun_MNIST_page}, which is supplied in values from 0 to 255, by scaling to the range $0.02-0.98$.  Perhaps a better representation could be chosen\cite{Fischer07cv} in a later test.  Each of the test characters is negated (subtracted from $1$) and is an input vector instance $\mathbf{x_k}$.  Classification output is a straightforward 1-NN search - essentially performing a brute lookup.  There is no training step or backpropagation.  

The weights are instantiated sequentially over $j$ to the 60,000 MNIST training characters.  In deference to Equation \ref{eq:_neuron_der_nd_w}, these weight vectors are transformed in preprocessing to the fourth power of the its values.  Each assignment iteration yields a candidate.  Overall, no activation function is used, and the winning output candidate is taken as the correct prediction for the test digit.\footnote{Note here that the winning candidate is the one with the lowest value, since we are looking for lowest weighted maximum discrepancy.}

I ran this scenario several times over differing values of the generalization parameter $p$.  The best result occurred when $p=-3$ which gives a test error of $\mathbf{3.04}\% $, with 9,696 of 10,000 correct, which is similar to other K-nearest-neighbors results\cite{LeCun_MNIST_page} with no preprocessing.   Since this test uses a 1-NN search (a slow, exhaustive lookup), it would be trivial to add another digit or character to the classification set - such as a decimal point or comma - by adding examples to the training set.  On the other hand, because there is no learning, there is also no generalization. 

\subsection{Inside-Outside Search Test Using Two Layers}
Using the same MNIST data, a human might might employ a "common sense" approach and say, \textit{For each digit, let's look through masks of the candidates and call it a match if the whole mask is solidly filled for the interior of the digit and if the exterior of the digit is solidly empty.}  Here I engineer a test where the interior is selected by the weights assigned to the values of the candidate (training digit).  

I again preprocessed the regular and copied inverted digits using a nonlinear transform\footnote{A dilation and erosion operator would also work to preprocess the data} to avoid the edge aliasing and intermediate values.  The two copies are then appended as one input vector.  The two layer Composition I XNOR (see Equation \ref{eq:me_logic_type_i}) is accomplished using $p=5$ and $p=-3$.  The winning candidate is selected by taking the geometric mean of the top 4 highest values for each digit.  The threshold value was set at $1/14$ in this test.   The result was 9112 correct out of 9784 test digits, giving a coverage of $97.8\%$ and a test error of $6.9\%$, but the test is a humanized approach.

\section{The Multiplet Definition}

It may be useful to modify the initial multiplet definition by replacing denominator power $p-1$ with $p-q$, so that a further generalized form is
\begin{equation}
\label{eq:hybrid_power_lehmer__neuron}
 b + m {({\sum { w_i x_i^p } / \sum { w_i x_i^{p-q}}})}^{\frac{1}{q}}
\end{equation}
which is a rewritten Gini mean \cite{bullen_2003,GOULD1984611}.  It operates as a quadratic mean when $p=2$, $q=2$. In this form, the curves of increasing $p$ become surfaces on the $p,q$ plane. 

 Many interesting papers were written early in the development of non-Euclidean neural networks\cite{DuchAD99}, to present a bridge between Radial Basis Function networks and standard networks\cite{LehtokangasS98}.  Other excellent papers have started a substantial thread with discussion of hyperbolic spaces\cite{GaneaBecHof2018}.
 
\subsection{Definition}
There is an opportunity here to drop the root term $1/q$ and to define the $jth$ neuron in a multiplet, having the same input vector instance $\mathbf{x}$ and membership selection weights $\mathbf{w}$, as 
\begin{equation}
\label{eq:hybrid_power_lehmer_naked}
\boxed{ M_{j}(\mathbf{x}) = b_j + m_j {({\sum { w_i x_i^{p_j} } / \sum { w_i x_i^{p_j-q_j}}})} }
\end{equation}
with $p$ and $q$ as generalization parameters, affine transform parameters $m_j$ and $b_j$, and $q$ as the overall degree (of $x_i$).

\begin{figure}[ht]

\begin{tikzpicture}
\label{tikzpicture:distcurves_pic_by_q}
\begin{axis}[
    title={},
    xlabel={Power Parameter $q$ },
    ylabel={Unweighted Power Multiplet},
    xmin=-2, xmax=7,
    ymin=0, ymax=1.5,
    legend pos=north west,
    ymajorgrids=true,
    grid style=dashed,
]
 
\addplot[
    color=blue,
    mark=square,
    ]
    coordinates {
(-2.0,1.235)(-1.5,1.172)(-1.0,1.111)(-0.5,1.054)(0.0,1.0)(0.5,0.9483)(1.0,0.8991)(1.5,0.852)(2.0,0.8066)(2.5,0.7624)(3.0,0.7185)(3.5,0.6734)(4.0,0.6247)(4.5,0.5684)(5.0,0.4985)(5.5,0.4087)(6.0,0.2991)(6.5,0.1858)(7.0,0.09571)
    };
    
\addplot[
    color=black,
    mark=diamond,
    ]
    coordinates {
(-2.0,1.323)(-1.5,1.238)(-1.0,1.156)(-0.5,1.076)(0.0,1.0)(0.5,0.9261)(1.0,0.8546)(1.5,0.7852)(2.0,0.7178)(2.5,0.6522)(3.0,0.5882)(3.5,0.5256)(4.0,0.4642)(4.5,0.404)(5.0,0.3447)(5.5,0.2859)(6.0,0.2275)(6.5,0.1694)(7.0,0.1137)
    };
    
\addplot[
    color=red,
    mark=triangle,
    ]
    coordinates {
(-2.0,1.315)(-1.5,1.23)(-1.0,1.149)(-0.5,1.073)(0.0,1.0)(0.5,0.931)(1.0,0.8655)(1.5,0.8035)(2.0,0.7447)(2.5,0.6891)(3.0,0.6364)(3.5,0.5867)(4.0,0.5396)(4.5,0.495)(5.0,0.4524)(5.5,0.4107)(6.0,0.3673)(6.5,0.3165)(7.0,0.2497)
    };
    \legend{0.1 0.1 0.2 0.3 0.9, 0.1 0.3 0.5 0.7 0.9, 0.1 0.7 0.8 0.9 0.9}
    \addplot[
    color=green,thick,dashed,
    mark=none,
    ]
    coordinates { (1.0,0.0)(1.0,1.4) };
\end{axis}
    \node [right,color=green] at (3.5,1.5) {$Lehmer$};

\end{tikzpicture}

\caption{Effect of Second Generalization Parameter $q$ on the Calculated Maximum $p=7$.  See also Figure \ref{figure:pq_surface_normdist}}
\label{figure:distcurves_by_q}
\end{figure}
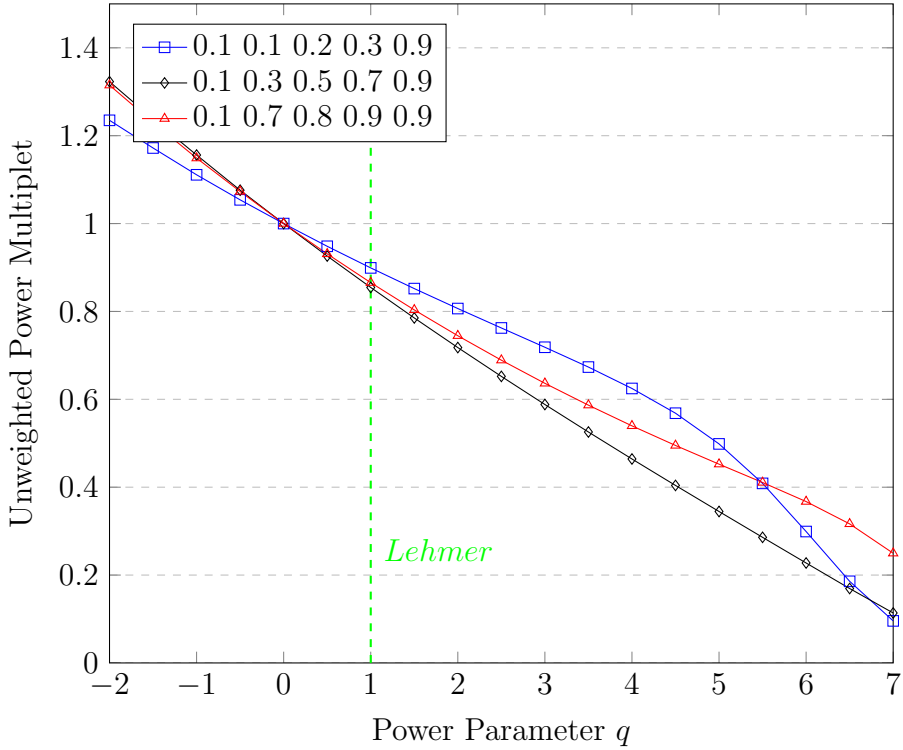


\begin{figure}[ht]
\centering\includegraphics[width=0.7\linewidth]{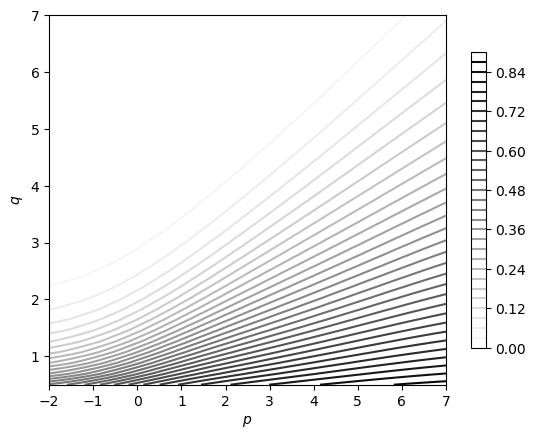}
\caption{The $p,q$ surface of a vector from a normal distribution, showing the tendency toward the minimum element value}
\label{figure:pq_surface_normdist}
\end{figure}

The total number of parameters $\phi_n$ in each neuron multiplet is now
\begin{equation}
\label{eq:multiplet_neuron_num_params}
  \phi_n = n + 4 \psi
\end{equation}
where $n$ is the number of input vector $\mathbf{x}$ elements, four is from the other parameters $b_j,m_j,p_j,q_j$ in each neuron, and the number of neurons in the multiplet is given by $\psi$.  

The effect of $q$ on the calculated maximum at $p=7$ may be seen in figure \ref{figure:distcurves_by_q}; note that as $q \rightarrow p$, the output declines toward the minimum\footnote{The calculated minimum - with $p$ negative - might be plotted on a log scale}. 

\subsection{Derivatives}

The derivative with respect to the weight $w_k$ is
\begin{equation}
\label{eq:pwr_neuron_derivative}
m \frac {x_k^p \sum { w_i x_i^{p-q} } - x_k^{p-q} \sum { w_i x_i^{p} }} {[{\sum { w_i x_i^{p-q}}]}^2}
\end{equation}
which can be rewritten in terms of the numerator sum $N=\sum { w_i x_i^p}$ and denominator $D=\sum { w_i x_i^{p-q}}$ as
\begin{equation}
\label{eq:pwr_neuron_der_nd}
\frac{\partial }{\partial w_k} ( b +  m { \frac{N}{D}}) = m \frac {D x_k^p - N x_k^{p-q}}  {D^2}
\end{equation}
similar the $p-1$ version.  For $w_i^L$, of course
\begin{equation}
\label{eq:pwr_neuron_der_nd_w2L}
\frac{\partial }{\partial w_k} ( b +  m { \frac{N}{D}}) = m L w_k^{L-1} \frac {D x_k^p - N x_k^{p-q}}  {D^2}
\end{equation}
where $w_i^L$ is the previously discussed function of the weights.  The derivative due to $p$ is
\begin{equation}
\label{eq:pwr_neuron_w_r_to_p}
\frac{\partial }{\partial p} ( b +  m { \frac{N}{D}}) = m {\frac {D \sum {w_i x_i^p ln(x_i) } - N \sum { w_i x_i^{p-q} ln(x_i)} } {D^2}}
\end{equation}
and with respect to $q$ it is 
\begin{equation}
\label{eq:pwr_neuron_w_r_to_q}
\frac{\partial }{\partial q} ( b +  m { \frac{N}{D}}) = m {\frac {  N \sum { w_i x_i^{p-q} ln(x_i)} } {D^2}}
\end{equation}
which requires calculation of the natural logarithm for each element in the input vector\footnote{Perhaps $ln(x_i)$ could be calculated concurrently with $x_i^p$}.

\section{On the Weighted Multiplet Perceptron Network}
In the weighted multiplet perceptron, the $w_i$ will adjust the aspect of the perceptron and the $m$ will adjust the threshold.  Let us begin by setting $q=2$.  When $p=2$, the perceptron has a circular or spherical shape.  When $p=3$, the perceptron has an elliptical or spheroid shape, but the surface may be discontinuous.  When $p=4$, a cuboidal shape results.  
Papers previously approaching this topic include centroid learning network concepts\cite{Lehtokangas00} and many others\cite{LuBBHN93}.

\begin{figure}[ht]
\centering
\begin{subfigure}
  \centering
  \includegraphics[width=.4\linewidth]{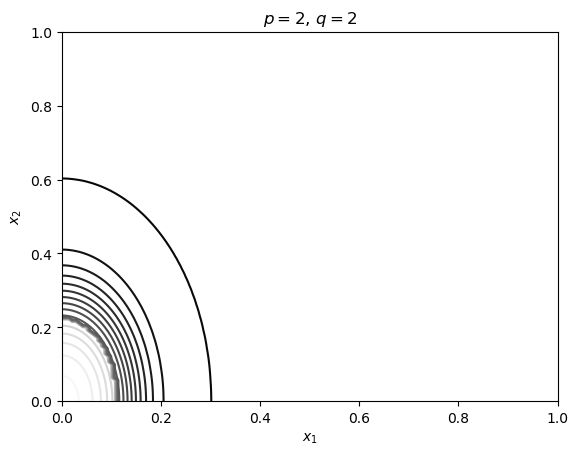}
\end{subfigure}
\begin{subfigure}
  \centering
  \includegraphics[width=.4\linewidth]{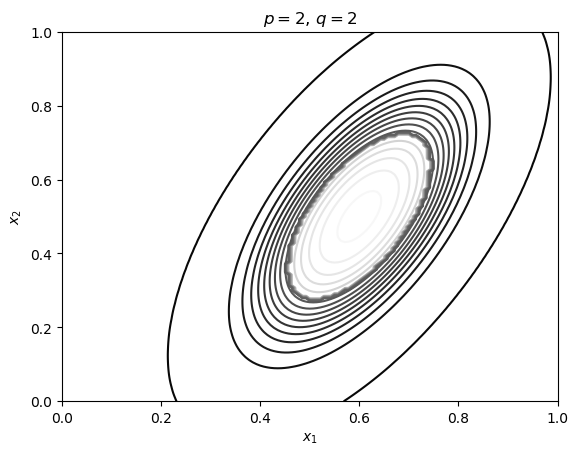}
\end{subfigure}
\centering
\begin{subfigure}
  \centering
  \includegraphics[width=.4\linewidth]{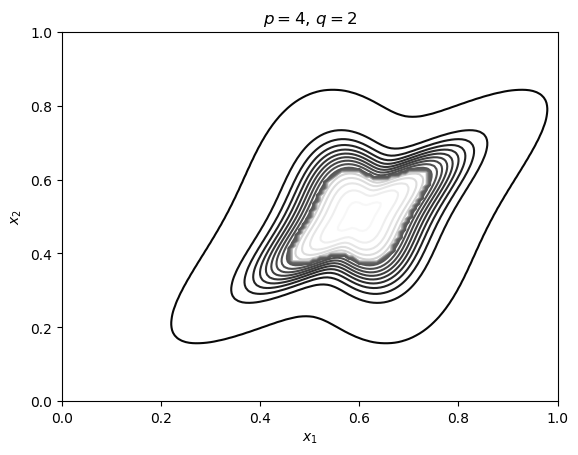}
\end{subfigure}
\begin{subfigure}
  \centering
  \includegraphics[width=.4\linewidth]{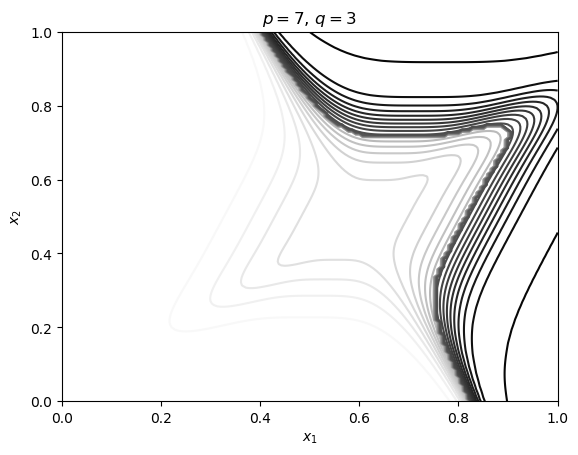}
\end{subfigure}
\caption{The Weighted Multiplet Perceptron Network, showing Different Cases of Localized Behavior} 
\label{figure:wtperceptron_cases}
\end{figure}

In figure \ref{figure:wtperceptron_cases}, the upper left panel depicts a circular decision boundary with $q=2$ and $p=2$ with $w_2=0.25$ and $m=50$.  The other panels show a two layer network where the first layer deploys a skew affine transformation.  In contrast to the perceptron examples in figure \ref{figure:perceptron_cases}, the multiplet perceptron can show localized behavior of the perceptron class boundary. With even generalization parameters, boundary enclosure can exist, which is an indication of potential superior capacity of the multiplet network.

\section{On Calculation of the Product of Vector Elements}

This section is perhaps a digression, but it is useful. The Lehmer mean case of $p=1/2$ is equivalent to the geometric mean \cite{ALZER1988c}, which of course uses the $nth$ root. For input vectors of size $n$ elements, can the expression
\begin{equation}
\label{eq:geom_aprox_mult_1_to_n}
{\sum_{i=1}^n {  x_i^p } / \sum {  x_i^{p-q}}}
\end{equation}
be approximately equal to element multiplication $\prod{x_i}$ for $q = n$?  Informally, the question posed here is \textit{"Can we set $q$ (and $p$) to compensate for the $nth$ root of the geometric mean and provide the product?"}  Let us try.

\subsection{On Conditions for Multiplication in One Layer}
For $n=1$, let $a=x_1$. Let me restate the pass-through property
\begin{equation}
\label{eq:x_passthrough}
    a^1 / a^0 = a
\end{equation}
in which the input to the layer passes through to the next layer when $q=1$.\footnote{One way to linearly  transform a layer is by letting $q=1$ and having one multiplet per element.}  We can easily calculate $a^2$ by setting $p=2$ and $q=2$ so that 
\begin{equation}
\label{eq:x_squared}
    a^2 / a^0 = a^2
\end{equation}
The same $a^2$ results if we let $p=1$
\begin{equation}
\label{eq:x_squared1}
    a^1 / a^{-1} = a / (1/a) = a^2
\end{equation}
or if $p=-3$
\begin{equation}
\label{eq:x_squared3}
    a^{-3} / a^{-5} =  a^2
\end{equation}
so that for one element (i.e. $n=1$), $q$ sets the degree of the result.

For $n=2$, let $a=x_1$ and $b=x_2$ and let $q=2$ with $p=1$, we have
\begin{equation}
\label{eq:geom_aprox_mult2}
  (a+b)/(a^{-1}+b^{-1}) = a b (a+b)/(b+a) = a b = x_1 x_2
\end{equation}
which is exactly $\prod{x_i}$ for any two $x_i$.  Note also that for $q=4$ and $p=2$, for two elements
\begin{equation}
\label{eq:geom_aprox_mult2_squared}
  (a^2+b^2)/(a^{-2}+b^{-2}) = a^2 b^2
\end{equation}
If we allow for $q=-2$ and $p=-1$, note that we have the inverse
\begin{equation}
\label{eq:geom_aprox_div2}
  (a^{-1}+b^{-1})/(a + b) = \frac{1}{a b}
\end{equation}
which is the exact inverse product of two scalar elements.  Division $a/b$ can occur in two layers, by
\begin{equation}
\label{eq:x_division}
    \frac{a^2 + 1/{ab}}{{a^2}^{-1} + {\frac{1}{ab}}^{-1}} = \frac{a^2(a^3b + 1)}{ab(a^3b+1)}= \frac{a}{b} = \frac{x_1}{x_2} 
\end{equation}
where $a^2$ and $1/ab$ are calculated (by different multiplets) in the first layer.  We will look to utilize this if possible.

Now consider $n=3$ and the positive reals. For three elements, the geometric mean takes the cube root and we want to use $q=3$. Let $r=p=3/2$
\begin{equation*}
(a^r + b^r + c^r)/ (a^{-r} + b^{-r} + c^{-r})  =(a^r b^r c^r (a^r + b^r + c^r))/(b^r c^r + a^r c^r + a^r b^r)
\end{equation*}
which is not the product $abc$.  However, if we require that $a=c$,  then delta from  $\prod{x_i} =a^2b$ is
\begin{equation}
 a^2 b - (2 a^r + b^r)/(2 a^{-r} +  b^{-r})
\end{equation}
which calculates in [0.01,1] as a generally flat surface about zero with a median of zero (within precision limits) and standard deviation of 0.008.  If we introduce weight terms into equation \ref{eq:geom_aprox_mult_1_to_n} to explore whether the weighted equation can perform the multiplication exactly, the reader can verify that when solved for a weight term, it is a trivial result in which $a$,$b$, and $c$ are required to be equal.
For $n=4$, $q=4$ and $p=2$, we use $a,b,c,d$ so that
\begin{multline}
\label{eq:geom_aprox_mult4}
(a^2 + b^2 + c^2 + d^2) / (a^{-2} + b^{-2} + c^{-2} + d^{-2}) \\
=(a^2 b^2 c^2 d^2 (a^2 + b^2 + c^2 + d^2))/(a^2 b^2 c^2 + a^2 b^2 d^2 + a^2 c^2 d^2 + b^2 c^2 d^2)
\end{multline}
which is not as tidy, but if we require $a=c$ and $b=d$, this reduces nicely to the product
\begin{equation}
    (2 a^2 + 2 b^2)/(2 a^{-2} + 2b^{-2}) = a^2 b^2 = a b c d
\end{equation}
exactly.  If we can require $a = c = d$, then the delta from $\prod{x_i} =a^3b$
\begin{equation}
a^3 b - (3 a^2 +  b^2) / (3 a^{-2} +  b^{-2})    
\end{equation}
presents another very flat surface about zero with median absolute error of $0.00057$ and standard deviation of $0.0157$.  However, this has median absolute percent error from the product $a b c d$ in (0,1] of $7.5\%$ - which seems good, but some of these products are off by an order of magnitude!

Regardless, I chose to further pursue this numerically, and I have calculated for a $n=7$ size vector with element values in [0.4,1]. The average absolute percent error is $10.6\%$, but the approximation can be off by as much as a factor of two - much more for if the values are allowed to approach zero.

\subsection{Exact Multiplication of Vector Elements in Multiple Layers}
Except for the two element case stated in Equation \ref{eq:geom_aprox_mult2}, the product of more than one input vector element cannot be reliably calculated in one layer.  However, the product $\prod{x_i}$ of the elements of a vector of even length $n$ can be exactly calculated in multiple layers by multiple neurons, if the weights are set exactly to construct a sort of binary tree.  

For example, in one multiplet neuron, the weights select $x_1$ and $x_2$ by $w_1=w_2=1.0$ and $w_3=w_4=0$, and in another multiplet in the same layer, the weights $w_i$ likewise select input elements $x_3$ and $x_4$.  Let all multiplets have a neuron in which $p=1$ and $q=2$, which have outputs that are fed into the next layer without activation.  

Similarly in the next layer, let a multiplet have the same behavior, selecting these two outputs with weights $w_i$.  The aggregate product $(x_1 x_2) (x_3 x_4)$ will be calculated exactly.  This, of course, as in a binary tree, requires $k$ layers where $2^k=n$.  In this simple example $k=2$ since $n=4$ elements in the input vector.  Moreover, it requires at least $n/2$ separate multiplets to coordinate weight parameters in the first layer alone - not likely to happen in a simple gradient descent system without constraints on sparsity.

\section{The Single-Element Power Series in Two Layers}
When $p = q$, the multiplet neuron expression (Equation \ref{eq:hybrid_power_lehmer_naked}) can be a monomial in $x_i$ of power $q$, which can be combined into a polynomial by the next layer.  

\begin{figure}

\begin{center}

\tikzset{every picture/.style={line width=0.75pt}} 

\begin{tikzpicture}[x=0.75pt,y=0.75pt,yscale=-1,xscale=1]

\draw   (271.65,99.74) .. controls (271.65,83.43) and (291.43,70.21) .. (315.83,70.21) .. controls (340.22,70.21) and (360,83.43) .. (360,99.74) .. controls (360,116.04) and (340.22,129.26) .. (315.83,129.26) .. controls (291.43,129.26) and (271.65,116.04) .. (271.65,99.74) -- cycle ;
\draw    (230,90) -- (274,91) ;
\draw   (470,130.48) .. controls (470,114.17) and (489.78,100.95) .. (514.17,100.95) .. controls (538.57,100.95) and (558.35,114.17) .. (558.35,130.48) .. controls (558.35,146.78) and (538.57,160) .. (514.17,160) .. controls (489.78,160) and (470,146.78) .. (470,130.48) -- cycle ;
\draw   (271.65,169.74) .. controls (271.65,153.43) and (291.43,140.21) .. (315.83,140.21) .. controls (340.22,140.21) and (360,153.43) .. (360,169.74) .. controls (360,186.04) and (340.22,199.26) .. (315.83,199.26) .. controls (291.43,199.26) and (271.65,186.04) .. (271.65,169.74) -- cycle ;
\draw    (229,166) -- (273,167) ;
\draw   (271.65,30.48) .. controls (271.65,14.17) and (291.43,0.95) .. (315.83,0.95) .. controls (340.22,0.95) and (360,14.17) .. (360,30.48) .. controls (360,46.78) and (340.22,60) .. (315.83,60) .. controls (291.43,60) and (271.65,46.78) .. (271.65,30.48) -- cycle ;
\draw    (230,20.74) -- (274,21.74) ;
\draw  [fill={rgb, 255:red, 0; green, 0; blue, 0 }  ,fill opacity=1 ] (314,217) .. controls (314,215.34) and (315.34,214) .. (317,214) .. controls (318.66,214) and (320,215.34) .. (320,217) .. controls (320,218.66) and (318.66,220) .. (317,220) .. controls (315.34,220) and (314,218.66) .. (314,217) -- cycle ;
\draw  [fill={rgb, 255:red, 0; green, 0; blue, 0 }  ,fill opacity=1 ] (314,253) .. controls (314,251.34) and (315.34,250) .. (317,250) .. controls (318.66,250) and (320,251.34) .. (320,253) .. controls (320,254.66) and (318.66,256) .. (317,256) .. controls (315.34,256) and (314,254.66) .. (314,253) -- cycle ;
\draw  [fill={rgb, 255:red, 0; green, 0; blue, 0 }  ,fill opacity=1 ] (314,237) .. controls (314,235.34) and (315.34,234) .. (317,234) .. controls (318.66,234) and (320,235.34) .. (320,237) .. controls (320,238.66) and (318.66,240) .. (317,240) .. controls (315.34,240) and (314,238.66) .. (314,237) -- cycle ;
\draw    (360,30.48) -- (480,110) ;
\draw    (360,99.74) -- (470,130.48) ;
\draw    (360,169.74) -- (470,140) ;

\draw (211,80.4) node [anchor=north west][inner sep=0.75pt]    {$x_{1}$};
\draw (291,80.4) node [anchor=north west][inner sep=0.75pt]    {$m=a_{1}$};
\draw (295,102.4) node [anchor=north west][inner sep=0.75pt]    {$q\ =1$};
\draw (291,150.4) node [anchor=north west][inner sep=0.75pt]    {$m=a_{2}$};
\draw (295,172.4) node [anchor=north west][inner sep=0.75pt]    {$q\ =2$};
\draw (210,156.4) node [anchor=north west][inner sep=0.75pt]    {$x_{1}$};
\draw (211,11.14) node [anchor=north west][inner sep=0.75pt]    {$x_{1}$};
\draw (291,11.14) node [anchor=north west][inner sep=0.75pt]    {$m=a_{0}$};
\draw (295,32.4) node [anchor=north west][inner sep=0.75pt]    {$q\ =0$};
\draw (491,132.4) node [anchor=north west][inner sep=0.75pt]    {$q=1$};
\draw (491,112.4) node [anchor=north west][inner sep=0.75pt]    {$w_{i} =1$};

\end{tikzpicture}

\end{center}

\caption{A truncated power series of one variable $x_1$ in two layers, where we set $p=q$ and $b=0$}
\label{figure:diagram_powser_multiplet}
\end{figure}
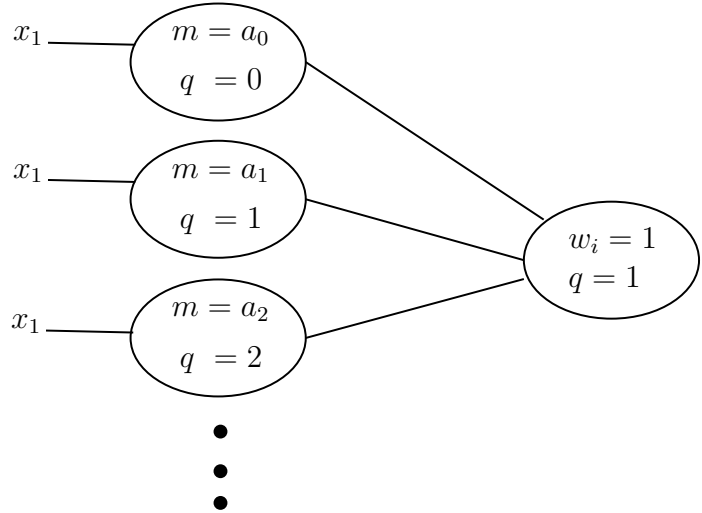

The power series in one variable $x_1$, stated generically as 
\begin{equation}
\label{eq:power_series0}
\sum {a_k * (x_1 - c)^k}
\end{equation}
can be constructed explicitly by a two-layer multiplet network.\footnote{Constant $c$ is assumed to have been subtracted in a previous layer}  Letting $a_k \rightarrow m_k$ and $w_i = 0$, except $w_1 = 1.0$, we have terms in the first layer
\begin{equation}
\label{eq:power_series_terms_Q}
(m_k * 1.0 x_1^k) / 1.0
\end{equation}
so that the power series sum, accomplished in layer two, is approximated by the chosen number of neurons 
\begin{equation}
\label{eq:power_series_approx4}
\sum {m_k * x_1^k} \simeq m_0 + m_1 * x_1 + m_2 * x_1^2 + m_3 * x_1^3 + m_4 * x_1^4
\end{equation}
where in this case we have the five multiplet neurons in layer one and the one neuron in layer two. See figure \ref{figure:diagram_powser_multiplet}. 

\subsection{The Power Series of More Elements in Two Layers}
If we leave in layer two $p=1$ so that summation occurs and set $w_1$ and $w_2$ to non-zero value, then a truncated power series in two elements and two layers (and two multiplets) is a construction that is linear.  The power series (where $p=q$) of two elements $x_1$ and $x_2$ with associated weights may be stated

\begin{equation}
\label{eq:power_ser_in_two_elems}
 \sum{a_0 (w_1 + w_2) + a_1 (w_1 x_1 + w_2 x_2) + a_2 ( w_1 x_1^2 + w_2 x_2^2 ) + ...} = \sum{w_i PS(x_i)}
\end{equation}
and $PS(x_i)$ is the power series of element $x_i$\footnote{The denominator here is the sum of all weights} so that \textit{the power series of a multi-element vector as expressed here is the sum of the power series of each element}.  Since $p=q$, the common denominator facilitates a linear relationship between power series of each input vector element.  Of course, this is not the same as a multivariate power series, where partials are taken and combined.

\subsection{Alternatives to Summation in Power Series}
If instead we set $p < 0$ in layer two, the summation in the truncated power series Equation \ref{eq:power_series0} would be replaced by a soft conjunction.  Of course, the standard power series with summation could be also be calculated within another neuron in the second layer.

\subsection{Some ubiquitous examples of power series in two layers}
The exponential function can be characterized by the power series
\begin{equation}
\label{eq:exp_by_power_series}
e^x = \sum_{k = 0}^{\infty} \frac{x^k}{k!} = 1 + x + \frac{x^2}{2} + \frac{x^3}{6} + \frac{x^4}{24} + \cdots
\end{equation}
which is well approximated for [0,1] by only these five terms in the equation.  The implies that the multiplet network could conceptually learn the parameters for $e^x$ within two layers, with only 5 neurons in the first layer, with $q=0,1,2,3,4$ and in the second layer $p=1$.   In general, multiplet networks of power series may be able to approximate in some interval
\begin{itemize}
    \item Trigonometric and Exponential functions
    \item The Geometric Series Result $a(1-r^n)/(1-r)$ or $a/(1-r)$
    \item The Log Expression $ln(1+x)$ 
    \item Derivatives and Special Products of Power Series
    \item Solutions of Differential Equations
\end{itemize}
given restrictions on the input, but further investigation is necessary to validate the number of terms and precision needed\footnote{Terms up to $x^6$ may be sufficient, depending on application} and other considerations.  Next, I present a short incursion into layer depth requirements.

\subsubsection{Layer Depth and the Softplus approximation}
The softplus function in one variable $ln(1+e^x)$ may be calculated in two approximations.  The first two layers may calculate the truncated power series approximation for $e^x = \xi$ and the next approximation of $ln(1+\xi)$ can occur in the next two layers.

However, if we take terms in Equation \ref{eq:exp_by_power_series} for $e^x = y$ and terms for the Taylor series of the natural log as
\begin{equation}
  \label{eq:ln_1_x_by_power_series}
  ln(1+y) =  y - 1/2 y^2 + 1/3 y^3 + \cdots
\end{equation}
we can directly input the first series into the second to obtain an approximation for softplus $ln(1+e^x)$ up to $\approx{0.3}$ as
\begin{equation}
    \label{eq:softplus_by_power_series}
\frac{5}{6} + x + x^2 + x^3 + \frac{19}{24}x^4 + \frac{1}{2}x^5 + \frac{2}{9}x^6 + \frac{5}{72}x^7 + \frac{1}{72}x^8 + \frac{1}{648}x^9 + \cdots
\end{equation}
which can be accomplished in two layers also.  The logistic function, formed from the exponential function and the geometric series, could be similarly reduced.

\begin{figure}[ht]
\centering\includegraphics[width=0.7\linewidth]{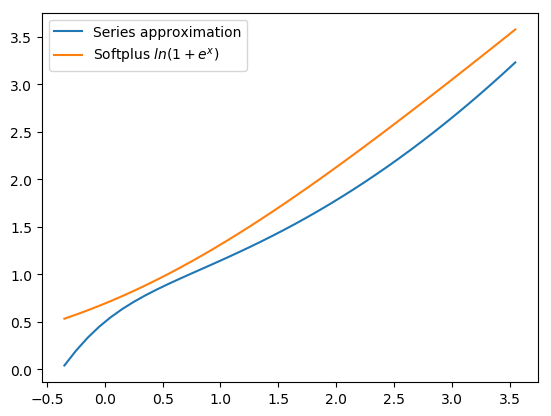}
\caption{Given by Equation \ref{eq:ln_1_x_Laurent_combo}, a heuristic series approximation of softplus.  Outside of the domain interval shown, it has a global minimum at $x=-1$ and is somewhat parabolic as $x \rightarrow \infty$}
\label{figure:xyplot_softplus2}
\end{figure}

A better approximation for softplus may be obtained if we decide to use some terms with negative exponents, such that $ln(1+e^x)$ is approximated by
\begin{equation}
    \label{eq:ln_1_x_Laurent_combo}
1/2 + \frac{1 + x + x^2/2}{4} + \frac{1}{4(1 + x + x^2/2)} - \frac{1}{2(1 + x + x^2/2)^2} + \dots
\end{equation}
where I have commandeered early terms from the Taylor series of $ln(x)$ at $1$ and some terms from the expansion of $ln(1+x)$ as $x \rightarrow \infty$.  Evidently, this requires four layers to implement, but only one output is needed from the second layer. Note in figure \ref{figure:xyplot_softplus2} that the accuracy is not high, since this is just for illustration, and the derivative will not be the same as that of the original softplus function.

\subsection{Series with Negative Exponents}
  
An instance in the multiplet neuron occurs when we set $p<0$ and $p=q$. This gives the neurons in the multiplet common denominators.  Although not as prevalent as power series, example expansions with negative exponents at $|z| \rightarrow \infty$ may include the natural log expression
\begin{equation}
    \label{eq:Puiseux_ser_ln_1_x}
    ln(1+z) = ln(z) + 1/z - \frac{1}{2z^2} + \frac{1}{3 z^3} + \cdots
\end{equation}
and the triangular difference
\begin{equation}
    \label{eq:Laurent_ser_sqrt_1_x2}
    -z + \sqrt{1+z^2} =\frac{1}{2z}-\frac{1}{8z^3}+\frac{1}{16z^5}-\cdots
\end{equation}
and the inverse relation
\begin{equation}
    \label{eq:series_1_x_m_1}
1/(z-1) = 1/z + (1/z)^2 + (1/z)^3 + (1/z)^4 + \cdots
\end{equation}
where $|z| > 1$ of course, as well as the truncated z-transform
\begin{equation}
    \label{eq:z_transform_m_pq}
    X(z) \simeq \sum{m_{j}(k) z^{-k}} 
\end{equation}
which may provide a measure of behavior of the $m_j$ across the multiplet.  However, this is no requirement at this time that $m_j$ be a continuous function or that neurons be contiguous in $p$ across the multiplet.

Series in a single variable $x_a$ in powers of $1/x_a$ have properties that can be a problem.  Terms of negative exponents may be needed in some circumstances, but we must determine what safeguards are necessary to assure safe computation.

\subsection{The Case of the Pad\'{e} Approximant in One Variable}
\begin{figure}

\begin{center}

\begin{tikzpicture}[x=0.75pt,y=0.75pt,yscale=-1,xscale=1]

\draw   (113.3,118.79) .. controls (113.3,102.48) and (133.08,89.26) .. (157.48,89.26) .. controls (181.87,89.26) and (201.65,102.48) .. (201.65,118.79) .. controls (201.65,135.09) and (181.87,148.31) .. (157.48,148.31) .. controls (133.08,148.31) and (113.3,135.09) .. (113.3,118.79) -- cycle ;
\draw    (71.65,109.05) -- (115.65,110.05) ;
\draw   (261.65,138.79) .. controls (261.65,122.48) and (281.43,109.26) .. (305.83,109.26) .. controls (330.22,109.26) and (350,122.48) .. (350,138.79) .. controls (350,155.09) and (330.22,168.31) .. (305.83,168.31) .. controls (281.43,168.31) and (261.65,155.09) .. (261.65,138.79) -- cycle ;
\draw   (113.3,188.79) .. controls (113.3,172.48) and (133.08,159.26) .. (157.48,159.26) .. controls (181.87,159.26) and (201.65,172.48) .. (201.65,188.79) .. controls (201.65,205.09) and (181.87,218.31) .. (157.48,218.31) .. controls (133.08,218.31) and (113.3,205.09) .. (113.3,188.79) -- cycle ;
\draw    (70.65,185.05) -- (114.65,186.05) ;
\draw   (113.3,49.52) .. controls (113.3,33.22) and (133.08,20) .. (157.48,20) .. controls (181.87,20) and (201.65,33.22) .. (201.65,49.52) .. controls (201.65,65.83) and (181.87,79.05) .. (157.48,79.05) .. controls (133.08,79.05) and (113.3,65.83) .. (113.3,49.52) -- cycle ;
\draw    (71.65,39.79) -- (115.65,40.79) ;
\draw    (201.65,49.52) -- (270,118.31) ;
\draw    (201.65,118.79) -- (260,138.31) ;
\draw    (201.65,188.79) -- (270,158.31) ;
\draw   (112.65,327.1) .. controls (112.65,310.79) and (132.43,297.57) .. (156.83,297.57) .. controls (181.22,297.57) and (201,310.79) .. (201,327.1) .. controls (201,343.4) and (181.22,356.62) .. (156.83,356.62) .. controls (132.43,356.62) and (112.65,343.4) .. (112.65,327.1) -- cycle ;
\draw    (71,317.36) -- (115,318.36) ;
\draw   (112.65,397.1) .. controls (112.65,380.79) and (132.43,367.57) .. (156.83,367.57) .. controls (181.22,367.57) and (201,380.79) .. (201,397.1) .. controls (201,413.4) and (181.22,426.62) .. (156.83,426.62) .. controls (132.43,426.62) and (112.65,413.4) .. (112.65,397.1) -- cycle ;
\draw    (70,393.36) -- (114,394.36) ;
\draw   (112.65,257.83) .. controls (112.65,241.53) and (132.43,228.31) .. (156.83,228.31) .. controls (181.22,228.31) and (201,241.53) .. (201,257.83) .. controls (201,274.14) and (181.22,287.36) .. (156.83,287.36) .. controls (132.43,287.36) and (112.65,274.14) .. (112.65,257.83) -- cycle ;
\draw    (71,248.1) -- (115,249.1) ;
\draw   (261.65,347.57) .. controls (261.65,331.27) and (281.43,318.05) .. (305.83,318.05) .. controls (330.22,318.05) and (350,331.27) .. (350,347.57) .. controls (350,363.88) and (330.22,377.1) .. (305.83,377.1) .. controls (281.43,377.1) and (261.65,363.88) .. (261.65,347.57) -- cycle ;
\draw    (201.65,258.31) -- (270,327.1) ;
\draw    (201.65,327.57) -- (260,347.1) ;
\draw    (201.65,397.57) -- (270,367.1) ;
\draw   (401.65,137.83) .. controls (401.65,121.53) and (421.43,108.31) .. (445.83,108.31) .. controls (470.22,108.31) and (490,121.53) .. (490,137.83) .. controls (490,154.14) and (470.22,167.36) .. (445.83,167.36) .. controls (421.43,167.36) and (401.65,154.14) .. (401.65,137.83) -- cycle ;
\draw    (350,138.79) -- (401.65,137.83) ;
\draw   (401.65,348.79) .. controls (401.65,332.48) and (421.43,319.26) .. (445.83,319.26) .. controls (470.22,319.26) and (490,332.48) .. (490,348.79) .. controls (490,365.09) and (470.22,378.31) .. (445.83,378.31) .. controls (421.43,378.31) and (401.65,365.09) .. (401.65,348.79) -- cycle ;
\draw    (340,158.31) -- (430,318.31) ;
\draw    (350,348.31) -- (400,348.31) ;
\draw   (511.65,238.79) .. controls (511.65,222.48) and (531.43,209.26) .. (555.83,209.26) .. controls (580.22,209.26) and (600,222.48) .. (600,238.79) .. controls (600,255.09) and (580.22,268.31) .. (555.83,268.31) .. controls (531.43,268.31) and (511.65,255.09) .. (511.65,238.79) -- cycle ;
\draw    (480,158.31) -- (520,218.31) ;
\draw    (480,328.31) -- (520,258.31) ;

\draw (50.65,96.05) node [anchor=north west][inner sep=0.75pt]    {$x_{1}$};
\draw (130.65,96.05) node [anchor=north west][inner sep=0.75pt]    {$m=a_{1}$};
\draw (134.65,118.05) node [anchor=north west][inner sep=0.75pt]    {$q\ =1$};
\draw (130.65,166.05) node [anchor=north west][inner sep=0.75pt]    {$m=a_{2}$};
\draw (134.65,188.05) node [anchor=north west][inner sep=0.75pt]    {$q\ =2$};
\draw (49.65,172.05) node [anchor=north west][inner sep=0.75pt]    {$x_{1}$};
\draw (50.65,26.79) node [anchor=north west][inner sep=0.75pt]    {$x_{1}$};
\draw (130.65,26.79) node [anchor=north west][inner sep=0.75pt]    {$m=a_{0}$};
\draw (134.65,48.05) node [anchor=north west][inner sep=0.75pt]    {$q\ =0$};
\draw (287,137.31) node [anchor=north west][inner sep=0.75pt]    {$q=1$};
\draw (280.65,117.31) node [anchor=north west][inner sep=0.75pt]    {$p=1$};
\draw (50,304.36) node [anchor=north west][inner sep=0.75pt]    {$x_{1}$};
\draw (130,304.36) node [anchor=north west][inner sep=0.75pt]    {$m=b_{1}$};
\draw (134,326.36) node [anchor=north west][inner sep=0.75pt]    {$q\ =1$};
\draw (130,374.36) node [anchor=north west][inner sep=0.75pt]    {$m=b_{2}$};
\draw (134,396.36) node [anchor=north west][inner sep=0.75pt]    {$q\ =2$};
\draw (49,380.36) node [anchor=north west][inner sep=0.75pt]    {$x_{1}$};
\draw (50,235.1) node [anchor=north west][inner sep=0.75pt]    {$x_{1}$};
\draw (130,235.1) node [anchor=north west][inner sep=0.75pt]    {$m=1$};
\draw (134,256.36) node [anchor=north west][inner sep=0.75pt]    {$q\ =0$};
\draw (287,347.31) node [anchor=north west][inner sep=0.75pt]    {$q=1$};
\draw (280.65,326.1) node [anchor=north west][inner sep=0.75pt]    {$p=1$};
\draw (427,137.31) node [anchor=north west][inner sep=0.75pt]    {$q=2$};
\draw (420.65,116.36) node [anchor=north west][inner sep=0.75pt]    {$p=1$};
\draw (419,347.31) node [anchor=north west][inner sep=0.75pt]    {$q=-2$};
\draw (419,327.31) node [anchor=north west][inner sep=0.75pt]    {$p=-1$};
\draw (537,237.31) node [anchor=north west][inner sep=0.75pt]    {$q=2$};
\draw (530.65,217.31) node [anchor=north west][inner sep=0.75pt]    {$p=1$};

\end{tikzpicture}

\end{center}

\caption{Diagram of a [2/2] Pad\'{e} Approximant in Four Layers, as in Equation \ref{eq:pade_degree2}, where the denominator terms are shown as the lowest 3 neurons in the input layer }
\label{fig:pade_diagram}

\end{figure}
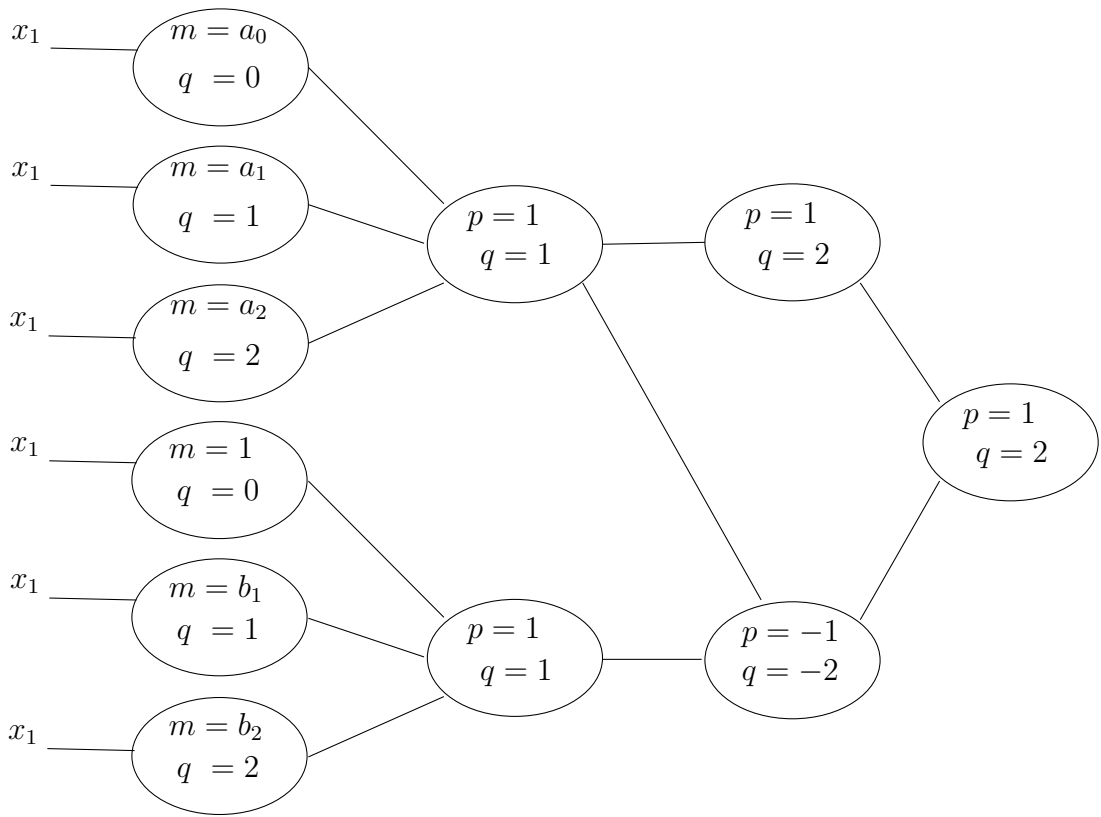

The Pad\'{e} Approximant of order [m/n] is the ratio of power series given by
\begin{equation}
\label{eq:pade_def}
    \frac{\sum_{j=0}^m a_j x^j}{1 + \sum_{k=1}^n b_k x^k} = \frac{a_0 + a_1 x + a_2 x^2 + \dots + a_m x^m}{1 + b_1 x + b_2 x^2 + \dots + b_n x^n}
\end{equation}
but let us consider the basic case of up to degree $2$ only.  Some layers are required to exactly calculate
\begin{equation}
\label{eq:pade_degree2}
    \frac{a_0 + a_1 x_1 + a_2 x_1^2 }{1 + b_1 x_1 + b_2 x_1^2 }
\end{equation}
but we already know the network can form a power series and a two element division.  Each term in the numerator and each term in the denominator may come from the same multiplet in the first layer.  The second layer would sum the numerator and the denominator in two separate multiplets, selecting from the six terms.   The third layer would then perform the square operation and the inverse two-term multiplication, as in Equation \ref{eq:x_division}.  See figure \ref{fig:pade_diagram}.  

The final multiply of the terms will be done in the fourth layer.   It is unlikely that this configuration would be something the network could learn without restrictions on connection sparsity in the latter layers. 

A recent paper\cite{MolinaSK19} introduced the Pad\'{e} activation unit, indicating that a parameterized approximant can increase predictive performance.  Their paper places an absolute value on the denominator in order to introduce stability.  Multiplet networks restrict the $w_i$ to positive values, but insuring a positive denominator could require constraints on other parameters of the multiplet neuron.

In the multiple element consideration, the multiplet power series in the numerator (and denominator) are formed by superposition of the individual variable power series.  There will be no $x_1^s x_2^t$ terms.  However, in the literature\cite{Chisholm73}, the approximant in a double power series has cross-terms between the variables.  The mathematical properties designed into the Canterbury approximant\cite{Chishom_McEwan74} cannot be assumed to hold within the multiplet network.

\section{Relating Input Vectors from Differing Distributions}
\begin{figure}[ht]
\centering
\begin{subfigure}
  \centering
  \includegraphics[width=.4\linewidth]{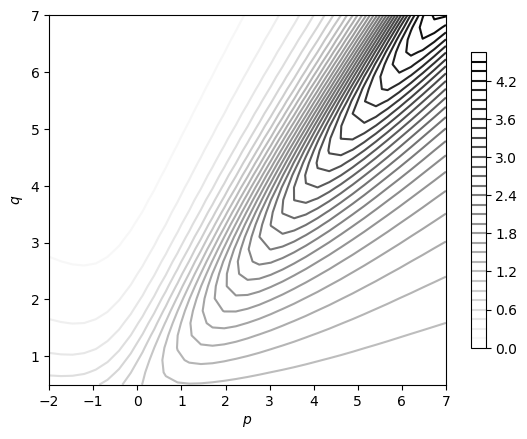}
\end{subfigure}
\begin{subfigure}
  \centering
  \includegraphics[width=.4\linewidth]{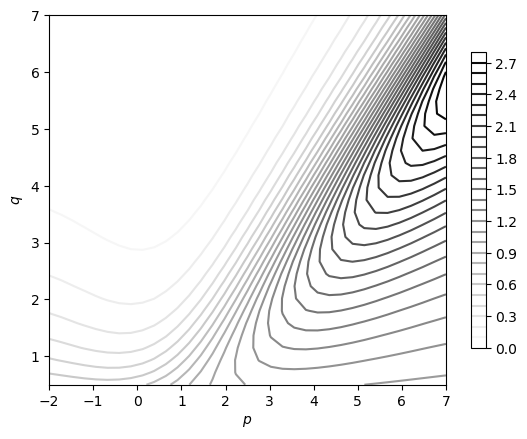}
\end{subfigure}
\caption{The ratio of a surface of an input vector from a left-skewed distribution to the surface generated from normal input, showing a ridge that rises with increasing $p$ and $q$. From a right-skewed distribution, the ridge appears further out}
\label{figure:pq_surface_ratio_lr_to_norm}
\end{figure}

I investigated the ratio of $p,q$ surfaces from two inputs.  The normalizing surface is the normal surface previously shown in figure \ref{figure:pq_surface_normdist}. For a vector from a somewhat left skewed distribution ( more high-valued elements ), the surface was generated, normalized, and plotted in figure \ref{figure:pq_surface_ratio_lr_to_norm}.  The surface is characterized by a somewhat linear ridge at an angle.  For a vector from a somewhat right skewed distribution, the normalized surface shows a similar ridge, but corresponding to higher $p$ value.  

These figures indicate that for a given value of $p$ and $q$, we can multiply a factor against the multiplet output to translate it to the represented output of a different distribution characteristic.  This factor would be taken from a selected prototype ratio surface generated from ideal distributions.

\section{Learning Rate Regularization Using the Case Slope Score}

One easy question in semi-supervised learning is to ask "Do we want the network to expressly pay attention to inputs that are somewhat homogeneous?"  Here I present a straightforward approach to instance evaluation.

\begin{figure}[ht]
\centering\includegraphics[width=0.7\linewidth]{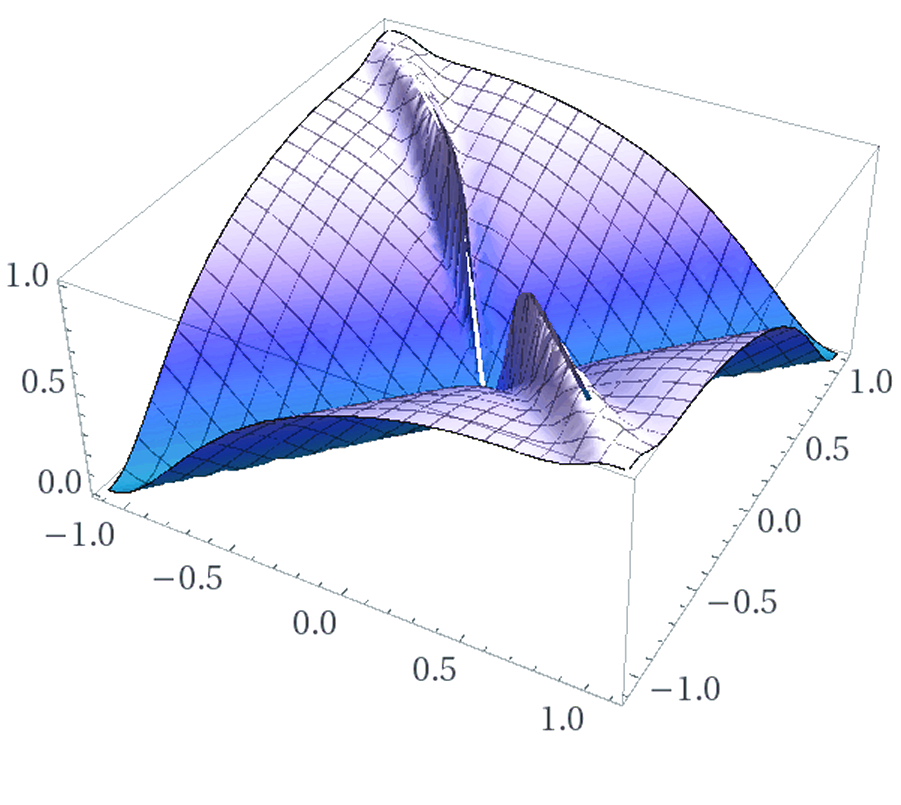}
\caption{Squashed Real Surface of Case Slope Score $\nu$ for Two Complex Elements $z_1$ and $z_2$, illustrating Lower $\nu$ Scores for Similar Input Values}
\label{figure:caseSlope3dGraphics}
\end{figure}

\subsection{The Mean Case Slope Score}
In manner analogous to calculating linear slope $\delta y/\delta x$, I choose (with $q=1$) two suitable values (one below and one above the arithmetic mean), such as $p_2=6$ and $p_1=-3$ to explicitly calculate the mean case difference (see figure $\ref{figure:distcurves}$) score:

\begin{equation}
\nu_0 = |{\sum { w_i z_i^6 } / \sum { w_i z_i^5 }} -  {\sum { w_i z_i^{-3} } / \sum { w_i z_i^{-4} }}| / 9
\label{eq:score_case_slope_score}
\end{equation}
where all operations are on complex numbers. The result $\nu_0$ is generally well behaved as long as $p_2$ and $p_1$ are even-odd pairs (and as long as the input vector is not perfectly anti-symmetric, e.g. -0.5,0.5,-0.5,0.5).  

The $\nu_0$ scores near zero indicate some homogeneity in the input vector $\mathbf{z}$.  Elements of $\mathbf{z}$ that are scattered produce higher scores.   Note that this also gives defined values for negative inputs as well.  Moreover, no T value assumption is required and other values that are near-congruent between 0 and T (e.g. 0.45,0.4) also produce a viable near-zero number.   

This equation is only a first order approximation to the Lehmer mean case curve slope.  Application of the score would involve some squashing operation, such as the hyperbolic tangent:
\begin{equation}
\label{eq:squashed_case_slope_score}
\nu = \tanh(|{\sum { w_i z_i^6 } / \sum { w_i z_i^5 }} -  {\sum { w_i z_i^{-3} } / \sum { w_i z_i^{-4} }}|)
\end{equation}
Shown in Table \ref{table:squash4input_css} are some input element values and case slope score.
\begin{table}

\centering
\begin{tabular}{l l l l l l}
\hline
$x_1$ & $x_2$ & $x_3$ & $x_4$ & $\nu$ \\
\hline
-0.78 & -0.9 & -0.85 & -0.75 & \textbf{0.04}\\
0.18 & 0.2 & 0.12 & 0.11 & \textbf{0.06}\\
-0.9 & -0.5 & 0.9 & 0.49 & \textbf{1}\\
1 & -0.9 & -0.9 & 0.11 & \textbf{1}\\
0.4 & 0.4 & 0.45 & 0.41 & \textbf{0.01}\\
\hline
\end{tabular}
\caption{The Case Slope Score $\nu$ of Selected Input Vector Elements, using Equation \ref{eq:squashed_case_slope_score}, where similar values give a $\nu$ near zero  }
\label{table:squash4input_css}
\end{table}

\subsection{Application}
See figure \ref{figure:caseSlope3dGraphics} for the mean case slope score depiction, showing a surface at near zero along the line in x-y plane from (-1,-1) to (1,1).  I believe we can use this result to directly dampen the $w_i$ learning rate for this neuron in a sort of fast adaptation novelty - assuming we want the membership selection weights to act to prefer similar input values.  Essentially, we are saying "I don't need to change these weights, since I somewhat like them the way they are (for now)".  This can change at each instance of the vector, or we can accumulate the score and use later.

During learning for a given layer, we want to use $\nu$ directly on $w_i$ learning rate $\lambda$, as in
\begin{equation}
  \label{eq:delta_is_nu_lambda}
    \Delta_i = \nu \lambda \delta_i
\end{equation}
where $\Delta_i$ is the amount added to a given $w_i$ during backpropagation.  Another implementation might use a linear function which has an offset added to $\nu$ or a softmax function of $\nu$, calculated across the layer.

\section{General Commments On the Multiplet Network Context}
In early networks, context was established by the use the layer offset. Even in a single layer, the weights in a traditional neural network may perform more advanced operations. For instance, the Savitzky-Golay filter is a transform that can emulate a running second-order least-squares regression smooth, and its properties are still being explored\cite{FLD2019}. 

In deep recurrent neural networks, the neurons may share weights through multiple layers\cite{Goodfellow-et-al-2016}. Convolutional neural networks\cite{LeCun2010c} perform transforms with prescribed weight sharing and max pooling. 

In multiplet networks, context is established through the use of the shared weights $w_i$ within the multiplet.  Each member of the multiplet is operating on the same inputs and can perform its own summarization transform.  

As introduced, the case slope score acts as a sort of homogeneity loss for the $w_i$ in point instance.  Other regularizations, especially those focused on the magnitude or norm, operate as part of a cost function.  So, the different parameters in the multiplet equation provide a way for segmentation for regularization, in that the $w_i$ can be regularized quickly, the $m$ and $b$ in standard time, and the $p$ and $q$ more slowly.

\section{Backpropagation}

Multiplet neurons can be regarded as regular neurons in backpropagation, with the exception that the multiplet will give a \textit{set of adjustments} to the parameters.  Whether a activation function layer is needed is open for debate.

The quantity of experiments that could be performed is beyond the scope of this paper.  However, as a quick application to the Iris dataset, I performed a comparison using two input elements.  The standard 4 layer network with 8 neurons in two hidden layers network took 4500 epochs to converge, with 11 classification outliers.  The 2 layer multiplet network needed only 12 parameters and converged within $1/10th$ of the time.

\section{Conclusion}

The multiplet network can select various means, perform sparse multiplications, provide interval-end semi-homogeneity estimation, and instantiate truncated power series. It can fit into the current learning stack or stand as an end-to-end system.  Moreover, the multiplet network provides opportunity to partition regularization strategy into entropy regularization using the case slope and regularization related to generating processes and characteristics using traditional techniques.  

I have avoided the topic of probability, except to allude to distributions in some figures. I have not made an unproven assertions regarding the multiple neuron, i.e. that it is analogous to a cumulative probability, or other claims. I have not made a specification on any terms that must be positive definite. Instead, I have tried to keep this introduction somewhat practical, focusing on empirical points that I estimate may convey some meaning.

The case slope score and other scores may be useful as a measurement for use with learning rate adaptation, but the dream is to one day develop a regularization-learning framework and associated cost function to let the network itself select regions and rules of regularization. Finally, I am hopeful that multiplets can lend new capability and capacity to artificial neural networks and that we can achieve more compatibility between human and machine.





\bibliographystyle{model1-num-names}
\bibliography{lmn_in_anns-27.bib}

\appendix

\section {On the behavior of complex vectors}
\label{appendx1}

In the calculation of exclusive-or, when a complex third input element is set to zero in the real part, the anticipated behavior of $\chi$ would be to give the same results as the two element input with just $x_1$ and $x_2$, since it is part of the summation.  However, if a small $\varepsilon$ is added to the imaginary component of each complex element, things are not so straightforward.

\begin{figure}[htbp]
\centering\includegraphics[width=0.7\linewidth]{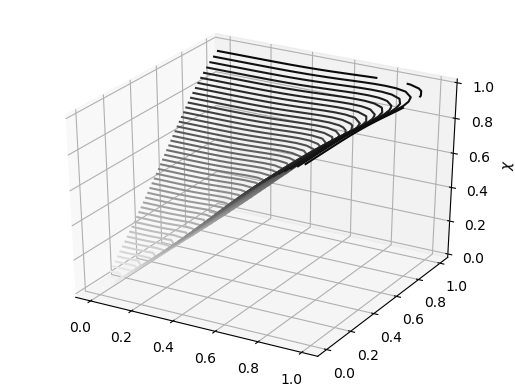}
\caption{When included in calculation of $\chi$, a complex third element $x_3$ with real part zero but non-zero imaginary part may still affect the calculation  }
\label{figure:xor_c_zero73}
\end{figure}
Even when the real part of $x_3$ is set to exactly zero - with the powers of $p$ at 7 and -3, $x_3$ still acts to affect $\chi$, as shown in Figure \ref{figure:xor_c_zero73}.

\section{On Potential Regularization and Constraints}
\label{section:someRegs}

Depending on the stated goal, here are logical, mathematical, or known properties of the generating process that can be invoked to assist machine learning.  If the network is to perform specific approximations, we may decide that parameters may have engineered constraints.  Some constraints in this section will address the factor $m$ in the multiplet.  Other constraints address $w_i$ - or both $m$ and $w_i$.  

These are constraints that the network designer would want enacted.  However, we must be careful in adopting dependent variables, since derivatives may be affected.

\subsection{By Mathematical Implementation}
As mentioned, values of $w_i$ are generally required to be positive.  As always, division by zero is to be avoided.  There may also be special constraints, such as positive denominator or numerator specification, or requirements for application within complex analysis.

\subsection{By Mathematical Parity}
If the generating process is known to be even or odd, generalization parameter $q$ might be required to be even or odd, at least in the first known layer.  Imposing this restriction on hidden layers is certainly another way to limit the behavior of the network, and it may be advantageous to select particular parity, by eliminating even powers or other method.

\subsection{By Independence or Sparsity Requirement}
At first, it may appear obvious that the requirement should be to insure that no neurons in the same multiplet have the same generalization parameters, assuming that the subsequent layer is using an arithmetic average.  Perhaps the network calculates the harmonic mean of neuron outputs in the next layer, and there is reason to let the network introduce a skew.  Therefore, the requirement might need to be relaxed, especially if the network is obtaining the $p$ and $q$ values by backpropagation or other learning.

If it is determined that the maximum variable independence is required in the problem, then careful engineering of the generalization parameters $p$ and $q$ must occur.   In particular, a required increment between values of $p$ may be set, e.g. 4, and all that is needed where $q=1$ is a three neuron multiplet with $p=5$, $p=1$, and $p=-3$.  

The case of the truncated power series is not going to have good independence between the terms. However, the specified independence requirement perhaps may be relaxed on certain layers, and required in others. 
The dot product case having negative coefficients, the exclusive-or, the range estimation, and the truncated power series are all implemented in multiplets in two layers.  The second layer often has less connections than the first. It may be necessary to impose a sparsity requirement on selected layers.

\subsection{By Logical Connective Construction}
As discussed, $m$ and $b$ must enable a logical complement for explicit calculation of the logical XOR in two layers.  Now, we have $q$ also that must be considered.  We must remember that derivatives must take into account any variables that are now functions of $p$ or $q$.

\subsection{To Select a Single Variable or Localize Attention}
To form a single variable power series exactly, all weights must be zero or nearly zero, other than the weight that selects the vector element.  A weight $w_k$ that is a function, such as
\begin{equation}
 \label{eq:single_var_weight_attention}
  w_k = w_i e^{-\alpha (i-k)^2}
\end{equation}
where $\alpha$ is some number (e.g.12.0), suppresses many other weights and emphasizes $w_k$.\footnote{Sparse connection, as used in standard convolution, could be accomplished by reducing the number of elements in the input.} If $k$ is not a hyperparameter but is learned by the network, we must consider the derivative with respect to $k$. 

To bring attention of the network to certain locally-related elements for learned convolution, a function similar to Equation \ref{eq:single_var_weight_attention} of more than one variable
\begin{equation}
 \label{eq:double_var_weight_attention}
  w_{jk} = w_i e^{-\alpha (i-j)^2} e^{-\alpha (i-k)^2}
\end{equation}
where  $\alpha$ is a suitable value.  Index variables $j$ and $k$ are related by metadata, such as height and width of an image.

\subsubsection{By Explicit Coefficient Properties}
Many useful expansions use factorials.  The mean of the Poisson distribution
\begin{equation}
    \label{eq:mean_of_poissondist}
    \sum_{k=0}^\infty k\frac{z^k}{k!} = z e^z
\end{equation}
is one example of the many series that have coefficients related to one another by index or factorial of the index variable.  This is obviously a restriction on the parameters in the network, in the case of power series terms.

If we have reason to coerce the network to construct an alternating series, we must decide how to embed a $(-1)^q$ type term within $m$, since $w_i$ are required to be positive.  Of course, this would affect the derivative with respect to $q$, since $m$ is now a function of $q$.  The Taylor series of sine is an example.

\subsection{To Perform Exact Multiplication}
As discussed in the previous section, if it is required that exact multiplication of elements occur, then several multiplets in each layer must coordinate weight parameters $w_i$.  The first multiplet will be required to have two $w_i$ parameters to be a constant 1.0 (with other $w_i$ at 0), and the next multiplet in the layer to have the subsequent two weights defined.  The next layer would handle the next level of binary lateral effect.

This can be conceptualized by a sort of boxcar function that is coupled to other boxcar functions.  
These functions are discontinuous, but there is another option - that the two selected elements need not be local.  Since multiplication is commutative, a multiplet can select, for example, $x_1$ and $x_3$ and the next multiplet can select $x_2$ and $x_4$.  Finally, a multiplet neuron has the potential to pass through an  element $x_i$ without modification - assuming there is no activation function that intervenes, and the product can be postponed.

\subsection{By Series Inversion}
The inversion of applied power series
\begin{equation}
    \label{eq:basic_aps_Y}
y = a_1 x + a_2 x^2 + a_3 x^3 + \dots   
\end{equation}
which is the problem of finding the coefficients $B_i$ in
\begin{equation}
    \label{eq:basic_aps_X}
x = B_1 y + B_2 y^2 + \dots
\end{equation}
is discussed in the literature\cite{ChernoffH47}. This is one way that would be possible to constrain the weights of the network - in that certain elements of a weight matrix which be required to be zero.

\subsection{By Recursion Relation}
The well known use of power series in solving linear differential equations can lead to the ubiquitous recursion relations between $a_q$ values.  I suggest that some constraints could be placed on the weights or other parameters between multiplet members, if the generating process is known to be related to a  differential equation.







\end{document}